\pdfoutput=1

\documentclass[11pt,breaklinks,colorlinks = true, linkcolor = blue, urlcolor = blue, citecolor = blue, anchorcolor = blue]{article}

\usepackage{ACL2023}
\usepackage{times}
\usepackage{latexsym}
\usepackage[T1]{fontenc}
\usepackage[utf8]{inputenc}
\usepackage{microtype}
\usepackage{comment}
\usepackage{tikz}
\usepackage{bbm}
\usepackage{booktabs}
\usepackage{soul}
\usepackage{bigstrut, tabularx, multirow, makecell, diagbox}
\usepackage{nicefrac}
\usepackage{multicol}
\usepackage{color}
\usepackage{colortbl,array,xcolor}
\usepackage{silence}
\usepackage{amssymb} %
\usepackage{arydshln} %
\usepackage{CJKutf8}
\usepackage{amsmath,amsfonts,amsthm}
\usepackage{hyperref}
\usepackage{graphicx} 
\usepackage{bm}
\usepackage{algorithm}    
\usepackage{algpseudocode}
\usepackage{ucs}
\usepackage{enumitem}
\usepackage{subcaption}
\usepackage{hhline}
\usepackage{geometry}
\usepackage{pgfplots}
\usepackage{overpic}
\pgfplotsset{compat=1.17} 
\usepackage{fancyhdr}
\usepackage{longtable}
\usepackage[hindi,main=english]{babel}

\newtheorem{proposition}{Proposition}

\title{Accelerating Transformer Inference for Translation via Parallel Decoding}

\author{Andrea Santilli\textsuperscript{1}, Silvio Severino\textsuperscript{1}, Emilian Postolache\textsuperscript{1},  Valentino Maiorca\textsuperscript{1}, \\ \textbf{Michele Mancusi\textsuperscript{1},
Riccardo Marin\textsuperscript{2,3}, Emanuele Rodolà\textsuperscript{1}}\\
  \textsuperscript{1}Sapienza University of Rome \hspace{0.24cm}
  \textsuperscript{2}University of Tübingen \hspace{0.2cm}\\
  \textsuperscript{3}Tübingen AI Center\\
  \texttt{santilli@di.uniroma1.it}}

\begin{document}
\maketitle

\begin{abstract}

Autoregressive decoding limits the efficiency of transformers for Machine Translation (MT). The community proposed specific network architectures and learning-based methods to solve this issue, which are expensive and require changes to the MT model, trading inference speed at the cost of the translation quality. In this paper, we propose to address the problem from the point of view of decoding algorithms, as a less explored but rather compelling direction. We propose to reframe the standard greedy autoregressive decoding of MT with a parallel formulation leveraging Jacobi and Gauss-Seidel fixed-point iteration methods for fast inference.
This formulation allows to speed up \textit{existing models} without training or modifications while \textit{retaining translation quality}. We present three parallel decoding algorithms and test them on different languages and models showing how the parallelization introduces a speedup up to 38\% w.r.t. the standard autoregressive decoding and nearly 2x when scaling the method on parallel resources. Finally, we introduce a decoding dependency graph visualizer (DDG\textit{viz}) that let us see how the model has learned the conditional dependence between tokens and inspect the decoding procedure.

\end{abstract}
\section{Introduction}
In recent years there have been dramatic improvements in Machine Translation (MT) \cite{edunov2018understanding, liu2020very} thanks to the transition to neural models and the advent of the Transformer architecture \cite{Vaswani:2017}. These models can produce high-quality translations while being extremely parallelizable during training.
However, Transformers are used sequentially at inference time, generating one token per time (i.e., sending each token as input for the next autoregressive iteration). This process of autoregressive inference hampers the efficiency of neural machine translation systems in terms of latency, limiting applications and portability.
Considering that these systems are extensively used in production multiple times to produce new translations (e.g., Google Translate\footnote{https://translate.google.com/}, DeepL Translator\footnote{https://www.deepl.com/}), even a minor speedup would be beneficial in the long run, especially if the translation is done on embedded devices.

To address this issue, the community proposed \textit{ad-hoc} trained models specific for parallel machine translation under the umbrella term of Non-Autoregressive Machine Translation models (NAT) \cite{gu:2018}. These models produce the translation in parallel but require (i) a complete reengineering of the MT system, (ii) extensive training resources and (iii) complex design choices like distillation from larger autoregressive models. These requirements are quite demanding and not easily satisfiable. For example, production systems are heavily optimized for hardware and software and even introducing a minimal modification requires non-trivial human effort \cite{wu2016google,kim-etal-2019-research}. Furthermore, training a new model from scratch is not always possible due to non-released training data or low-resource languages having few or lacking parallel corpora.

\begin{figure*}[!ht]
\begin{center}
\includegraphics[width=0.9\linewidth]{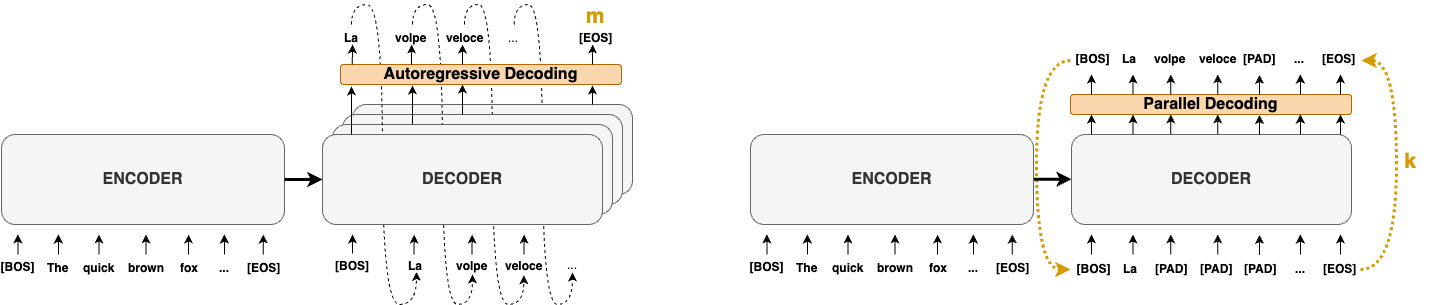}
\caption{\textbf{On the left}, the classical Autoregressive Decoding for MT. The target sentence is produced token-by-token sequentially, sending the {partial} result as input for the next autoregressive iteration up to the length $m$ of the target. \textbf{On the right} Parallel Decoding proposed in this paper. This method changes only the \textit{decoding algorithm} (orange block) and is usable on top of any autoregressive model without modifications.
Parallel Decoding algorithms resolve the whole sentence or a block of \textit{b} tokens in parallel: initial tokens (PAD tokens) are gradually refined with $k$ steps until a stopping condition is reached. Crucially, $k \leqslant m$ with quality guarantees and overall decoding speedups.}
\label{fig:teaser}
\end{center}
\vspace{-0.5cm}
\end{figure*}

In this paper, we propose to address the problem of parallel machine translation with an orthogonal approach consisting in novel decoding algorithms that work in parallel and can be used on top of \textit{existing autoregressive models} for MT. We overcome previous limitations with a flexible and generic method that does not require any modification to the model or costly retraining.
Specifically, inspired by previous successes in speeding up feedforward computation for image generation \cite{song:2021}, we reframe the greedy autoregressive decoding for MT as a system of nonlinear equations solvable in parallel. This simple formulation speeds up the decoding procedure by using fixed-point iteration methods like Jacobi and Gauss-Seidel while having mathematical guarantees on the quality of the translation. A high-level description of the method is available in (Fig. \ref{fig:teaser}).
Our contributions can be summarized as the following:
\begin{itemize}
    \item We reframe the standard greedy autoregressive decoding procedure in MT with a parallel formulation, introducing three parallel decoding algorithms (PJ, PGJ, HGJ) and a stopping condition that preserves translation quality.
    \item We perform extensive experiments with different transformer sizes (base and large) and datasets, showing speedups up to 38\% in time, obtaining a nearly ${2\boldsymbol{\times}}$ speedup when scaling the model on parallel resources while preserving quality.
    To the best of our knowledge, this is one of the first studies to introduce a speedup in multilingual machine translation.
    \item We introduce a decoding dependency graph visualizer (DDG\textit{viz}) to inspect the learned tokens' conditional dependence and when parallel decoding is effective.
\end{itemize}
All the code is publicly released\footnote{\url{https://github.com/teelinsan/parallel-decoding}}.

\section{Related Work}
\label{sec:related}

\begin{figure*}[!ht]
\begin{center}
        \begin{overpic}
        [trim=0cm 0cm 0cm 0cm,clip,width=0.99\linewidth]{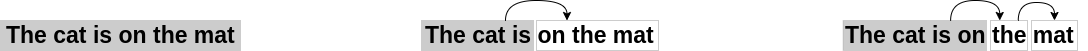}
        \put(10, -2.5){\textbf{\footnotesize{PJ}}}
        \put(48, -2.5){\textbf{\footnotesize{PGJ}}}
        \put(88, -2.5){\textbf{\footnotesize{HGJ}}}
        \end{overpic}
        \vspace{0.2cm}
        
\caption{\textbf{Parallel Decoding algorithms}: \textbf{PJ} resolves the whole sequence in parallel iteratively. \textbf{PGJ} resolves blocks in parallel; once a block is finished, it moves on to the next one and decodes it again in parallel (in figure $b= 3$). \textbf{HGJ} decodes the sentence in parallel as PGJ up to a certain length $h$; afterwards, it goes autoregressively until [EOS] token is generated. Decoding actually happens in sub-word tokens (not depicted here).}
\label{fig:algover}
\end{center}

\vspace{-0.5cm}
\end{figure*}

\citet{gu:2018} first introduced Non-Autoregressive Translation models (NAT) as ad-hoc trained models capable of producing the translation all at once in parallel. With NATs, it is possible to consistently reduce the latency and speed up the translation at the expense of a slightly worse translation quality due to the multimodality problem (i.e., we lose the dependency between tokens in the target output). Finding a tradeoff between translation quality and speed is an active research direction, with current methods trying to fill the gap in terms of translation quality \cite{Geng:2021, Savinov:2022}. Nevertheless, all proposed NAT models are learning-based and require different tricks to reach the quality of autoregressive models \cite{Gu:2021}. The most common is the sequence-level knowledge distillation of large autoregressive models into parallel models \cite{kim2016sequence}.
Other approaches include defining alternative training objectives \cite{Ghazvininejad:2020a, Saharia:2020, Du:2021, Huang:2021}, architectures that model dependencies between output sentence tokens \cite{Ghazvininejad:2019, Qian:2020, Song:2021:alignart, Gu:2021, song2022switchglat} or multi-iteration methods \cite{Ghazvininejad:2020b, Kasai:2020, Hao:2021, Geng:2021, Savinov:2022, Huang:2022, GAD:2022} that apply iterative refinements to a translation, trading some speed for greater quality.
In our approach, we also employ iterative refinements of solutions to non-linear equations, but \emph{we do not perform any training or modification to the model}.
Other works that require retraining or modifications to the model add additional decoding heads \cite{Stern:2018} or use shallow decoders \cite{kasai2021deep}. We refer the reader to \citet{xiao2022survey} for a thorough survey on NAT methods. Further orthogonal approaches use  specialized hardware (TPU) with low-precision calculations \cite{wu2016google} or software optimizations \cite{kim-etal-2019-research}.
In the context of Grammatical Error Correction, \citet{Sun:2020} recently proposed aggressive parallel decoding, assuming that the model output is similar to the input.
More recently, inspiring our work, \citet{song:2021} showed that it is possible to parallelize feedforward computations by thinking of them as a system of non-linear equations. 
They parallelized the backpropagation of RNNs, feedforward layers and autoregressive generative models on images.
We extend the approach defined on dense pixel prediction to the discrete conditional token generation in MT.
While this work was under submission and anonymity period, \citet{leviathan2022fast}, \citet{chen2023accelerating} and \citet{kim2023big} concurrently proposed decoding approaches that speed up inference of a large transformer model by using another smaller model to draft tokens. Compared to these approaches our method requires just an existing autoregressive model (no matter the size) and mathematically guarantees the output quality.
In the next Section we describe the method.

\section{Method}
In this Section, we introduce notations, develop the theory behind Parallel Decoding, present three algorithms (Fig. \ref{fig:algover}), and discuss the initialization and stopping conditions for the proposed approaches.

\subsection{Notation}
The goal of MT is to translate a sentence $\mathbf{x}$ in a source language (e.g., Italian) with its translation $\mathbf{y}$ in the target language (e.g., English). Source and target sentences are generally tokenized in words or subwords \cite{DBLP:journals/corr/abs-1808-06226, DBLP:conf/icassp/SchusterN12,sennrich-etal-2016-neural, kudo-2018-subword}; here, we use the subfix notation $\mathbf{x}=(x_1, \dots, x_n)$ and $\mathbf{y}=(y_1, \dots, y_m)$ to indicate specific tokens in the sequence. We also use the notation $\mathbf{x}_{1:n}$ to indicate a slice of a sequence as a shorthand of $\mathbf{x}=(x_1, \dots, x_n)$.
From a probabilistic perspective, an MT model estimates $p_\theta(\mathbf{y}\mid\mathbf{x})$. Once an MT model has been trained, the inference phase is traditionally performed by sampling tokens from the model probability conditioned on the input sequence $\mathbf{x}$ and previously generated tokens $(y_1, \dots, y_{i-1})$: 
\begin{align}
    p_\theta\left(y_{i} \mid y_{1}, \ldots, y_{i-1}, \mathbf{x}\right).
\label{eq:autoregressive}
\end{align}
Different sampling strategies are employed (e.g., Greedy, Top-K, Top-p \cite{JMLR:v21:19-985,Holtzman2020The}) alongside search strategies that estimate the total conditional probability (e.g., Greedy search, Beam search \cite{reddy}). The most straightforward strategy, Greedy Search, selects the element $y_i$ of a sequence with:
\begin{align}
\label{eq:greedy}
    y_{i}=\arg\max p_\theta(y_{i} \mid \mathbf{y}_{1:i-1}, \mathbf{x}).
\end{align}
Given the formalization above, a standard autoregressive setting runs $m$ inference steps \textit{sequentially} to generate an output sequence of $m$ elements.

\paragraph{Parallel Decoding.}
Given Equation \eqref{eq:greedy}, it is possible to write the greedy decoding procedure on all tokens as:
\begin{align}
\left\{\begin{array}{l}
y_{1}=\arg\max p_\theta(y_{1} \mid \mathbf{x}) \\
y_{2}=\arg\max p_\theta(y_{2} \mid y_{1}, \mathbf{x}) \\
\vdots \\
y_{m}=\arg\max p_\theta(y_{m} \mid \mathbf{y}_{1:m-1}, \mathbf{x}) \\
\end{array}\right.
\label{eq1}
\end{align}
Defining $f(y_{i}, \mathbf{y}_{1:i-1}, \mathbf{x}) = y_{i} - \arg\max p_\theta(y_{i} \mid \mathbf{y}_{1:i-1}, \mathbf{x})$ , we can rewrite the system of Equations \eqref{eq1} as:
\begin{align}
\left\{\begin{array}{l}
f(y_{1}, \mathbf{x}) = 0 \\
f(y_{2}, y_{1}, \mathbf{x}) = 0 \\
\vdots \\
f(y_{m}, \mathbf{y}_{1:m-1}, \mathbf{x})= 0 \\
\end{array}\right.
\label{eq2}
\end{align}
This system has $m$ non-linear equations (each equation employ a neural network) with $m$ variables.

\subsection{Parallel Decoding Algorithms}
\label{sect:paralledec}
The autoregressive decoding implicitly solves the system of Equations \eqref{eq2} by substitution, i.e., given the \textsc{[BOS]} token and the input sentence $x$, it solves equations from first to last, progressively replacing the resolved variables. In this paper, we rely on Jacobi and Gauss-Seidel (GS) fixed-point iteration methods \cite{ortega1970iterative} to solve in parallel system \eqref{eq2} until a stopping condition is reached.
This formulation is particularly flexible and has several advantages: Firstly, it is completely agnostic to the underlying MT model used; Secondly, it can be analyzed with analytical tools and has guarantees of convergence to the exact solution for system \eqref{eq2}; Thirdly, it can be potentially extended by drawing from the numerical methods literature for non-linear equations solving methods \cite{saad2003iterative}.
We see that, with the proper stopping condition, it is possible to have quality guarantees over the output.
We present here three algorithms (PJ, PGJ, HGJ) that leverage these fixed-point iteration methods to speedup decoding in MT.
\paragraph{Parallel Jacobi (PJ) Decoding.}
First, we propose Algorithm \ref{alg:pj}. This algorithm works by initializing a draft translation for the whole target sentence and then iteratively translating the whole sentence in parallel until the stopping condition is triggered. This is equivalent to solving system \eqref{eq2} with Jacobi, hence the name of the method.
\paragraph{Parallel GS-Jacobi (PGJ) Decoding.}
Decoding the whole target sentence in parallel may introduce difficulties in inferring long dependencies between tokens since the underlying model is trained to model the conditional distribution of a token given the previous tokens. In general, we observed that shorter dependencies are easily predicted since decoding happens at the sub-word level, and the model can decode sub-word unities in parallel rather than the whole sentence. To this end, we propose Algorithm \ref{alg:pgj}, called GS-Jacobi, that splits the sentence into contiguous $b$-dimensional blocks. Starting from the first one, it decodes in parallel all its elements. Once a block is finished or the stopping condition within the block is triggered, the algorithm performs a sequential (Gauss-Seidel) step and proceeds with (Jacobi) decoding on the next one.

\paragraph{Hybrid GS-Jacobi (HGJ) Decoding.}
Algorithms \ref{alg:pj} and \ref{alg:pgj} assume to know beforehand the number of equations $m$ (i.e., the target length). This is not usually the case for MT, where the model dynamically controls the length through the emission of a special end-of-sentence token \textsc{[EOS]}. To overcome this issue, we propose a flexible Hybrid Algorithm \ref{alg:phgj} that mixes PGJ computations with standard autoregressive decoding. This algorithm performs parallel  GS-Jacobi decoding up to a certain prefixed length $h$. If the \textsc{[EOS]} token is generated within a block, then the algorithm stops, returning the translation up to \textsc{[EOS]}. Otherwise, the algorithm concludes the translation by reaching the \textsc{[EOS]} token with standard autoregressive decoding. In this case, the length $h$ regulates the trade-off between parallel and sequential computation,
limiting the waste of resources beyond \textsc{[EOS]}.
\subsection{Initialization and Stopping}

\begin{algorithm}[t]
\footnotesize
\caption{Parallel Jacobi Decoding\label{alg:pj}}
\textbf{Input:} $\mathbf{x}=(x_1, \dots, x_n)$, $p_\theta$\\
\textbf{Output:} $\mathbf{y}=(y_1, \dots, y_m)$
\begin{algorithmic}[1]
\State $\mathbf{y} \gets \textsc{InitT}(\mathbf{x})$
\State $m \gets len(\mathbf{y})$
\For{$i=1$ to $m$}
\State $\mathbf{o} \gets copy(\mathbf{y}_{1:m})$
\State ${\mathbf{y}_{1:m}} \gets \arg \max(p_\theta({\mathbf{y}_{1:m}}| \mathbf{y}_{1:m}, \mathbf{x}))$
\State $stop \gets \textsc{StopC}(\mathbf{o}, \mathbf{y}_{1:m})$
\If{$stop$}
\State{break}
\EndIf
\EndFor\\
\Return $\mathbf{y}$

\end{algorithmic}

\end{algorithm}

Our algorithms share two components: the \textit{initialization procedure} and the \textit{stopping condition}.
\paragraph{Initialization $\textsc{InitT}(\textbf{x})$.}
The initialization procedure is a function that inputs the source sentence and produces an initial draft translation as output. In this paper we experimented with a simple initialization procedure that initialize the translation with all \textsc{[PAD]} tokens. This choice is fast and doesn't depend on the underlying MT model.
We leave as future work the research of different initialization procedures to further speedup the decoding.

\paragraph{Stopping Condition $\textsc{StopC}(\textbf{y}^{k-1}, \textbf{y}^{k})$.}
The stopping condition is a function that takes as input the previous-iteration sentence $\textbf{y}^{k-1}$ and the current-iteration sentence $\textbf{y}^{k}$ and decides whether to stop the algorithm or not. This function is crucial since it regulates the trade-off between speedup and translation quality. In this paper we introduce as stopping condition for MT:
\begin{align}
  \mathbf{y}^{k-1} - \mathbf{y}^{k} = \mathbf{0} 
 \label{eq:stopping}
\end{align}
i.e., the sentence from the previous step has not changed. This stop condition allows for preserving quality and quickening  translations simultaneously.

\begin{table*}[t]
\centering
\scriptsize
\begin{tabular}{l|cc|cc|cc|cc}
\multirow{2}{*}{\textbf{Decoding Algorithm}}&
\multicolumn{2}{c}{\textbf{en$\rightarrow$de}} &  
\multicolumn{2}{c|}{\textbf{de$\rightarrow$en}}  & 
\multicolumn{2}{c}{\textbf{en$\rightarrow$ro}} &  
\multicolumn{2}{c}{\textbf{ro$\rightarrow$en}} \\
 &  Speed\ & BLEU& Speed & BLEU & Speed & BLEU & Speed & BLEU \\
\hline
\textbf{Opus} & & & & & & & &\\
Greedy Autoregressive & $1.00\times$ & 28.24&$1.00\times$& 33.10 & $1.00\times$ & 27.41 & $1.00\times$ &  37.01\\
Beam Search (beam = 5) & $0.71\times$ & {28.68} & $0.72\times$ & 33.92 & $0.70\times$ &  {27.61} & $0.72\times$ &  {37.84} \\
\rowcolor[gray]{.90} PJ Decoding & $0.73\times$& 28.24 & $0.75\times$& 33.10 & $0.66\times$ &27.41 & $0.66\times$& 37.01\\
\rowcolor[gray]{.90} PGJ Decoding (b = 5) & $1.28\times$& 28.24 &$1.32\times$&33.10  & $1.33\times$ & 27.41 & $1.29\times$ & 37.01\\
\rowcolor[gray]{.90} PGJ Decoding (b = 3) & $\mathbf{1.34\times}$& 28.24 &$\mathbf{1.37\times}$&33.10 & $\mathbf{1.38\times}$ & 27.41 & $\mathbf{1.35\times}$ & 37.01\\
\rowcolor[gray]{.90} HGJ Decoding (b = 3) & $\mathbf{1.34\times}$ & 28.24& $\mathbf{1.37\times}$ & {33.10} &$\mathbf{1.38\times}$ &27.41 &$\mathbf{1.35\times}$ & 37.01 \\ 
\hline
\textbf{MBart50} & & & & & & & &\\
 Greedy Autoregressive & $1.00\times$ & 23.97&$1.00\times$& 31.58 & $1.00\times$ & 24.99 & $1.00\times$ &  34.77\\
Beam Search (beam = 5) & $0.76\times$ & {24.93} & $0.77\times$ & 32.61 & $0.77\times$ &  {25.31} & $0.76\times$ &  {35.16} \\
\rowcolor[gray]{.90} PJ Decoding & $0.88\times$& 23.97 & $0.88\times$& 31.58 & $0.86\times$ &24.99 & $0.85\times$& 34.77\\
\rowcolor[gray]{.90} PGJ Decoding (b = 5) & $0.98\times$& 23.97 &$0.98\times$&31.58 & $0.97\times$ & 24.99 & $0.99\times$ & 34.77\\
\rowcolor[gray]{.90} PGJ Decoding (b = 3) & $\mathbf{1.06\times}$& 23.97 &$\mathbf{1.08\times}$&31.58 & $\mathbf{1.03\times}$ & 24.99 & $\mathbf{1.04\times}$ & 34.77\\
\rowcolor[gray]{.90} HGJ Decoding (b = 3) & $1.05\times$ & 23.97& $1.07\times$ & 31.58 &$1.01\times$ &24.99 &$1.02\times$ & 34.77 \\ 
\hline
\end{tabular}
\label{en-de-ro_result}
\caption{Comparison of parallel decoding algorithms (highlighted in grey) with sequential decoding using Opus (CPU) and MBart50 (GPU) on WMT14 and WMT16. Speed is measured in time w.r.t. the autoregressive baseline.}
\label{tab:1}
\end{table*}

\begin{table*}[ht]
    \scriptsize
    \centering
    
        \begin{tabular}{llllllllllll}
        
        \multicolumn{2}{c}{} & \multicolumn{2}{c}{\textbf{WMT17}} & \multicolumn{2}{c}{\textbf{IITB}} & \multicolumn{2}{c}{\textbf{IWSLT15}}& \multicolumn{4}{c}{\textbf{FLORES}}\\
        \multicolumn{2}{c}{} & \multicolumn{2}{c}{\textbf{En-Fi}} & \multicolumn{2}{c}{\textbf{En-Hi}} & \multicolumn{2}{c}{\textbf{En-Vi}} &  \multicolumn{2}{c}{\textbf{En-It}} & \multicolumn{2}{c}{\textbf{En-Fr}} \\
        \cmidrule(lr){3-4} \cmidrule(lr){5-6} \cmidrule(lr){7-8} \cmidrule(lr){9-10} \cmidrule(lr){11-12}
        \textbf{Dec. Algorithm} & \textbf{Speed} &  $\text{ }\leftarrow$ & $ \text{ }\rightarrow$ & $ \text{ }\leftarrow$ & $ \text{ }\rightarrow$ & $ \text{ }\leftarrow$ & $ \text{ }\rightarrow$ & $ \text{ }\leftarrow$ & $ \text{ }\rightarrow$ & $ \text{ }\leftarrow$ & $ \text{ }\rightarrow$ \\ \midrule
         \multirow{2}{*}{{PJ}} & Iters & 1.04$\times$ & 1.04$\times$ & 1.04$\times$ & 1.04 $\times$& 1.06$\times$ & 1.03$\times$ & 1.02$\times$ & 1.04$\times$ & 1.03$\times$ & 1.03$\times$ \\ 
                                      & Time  & 0.86$\times$ & 0.88$\times$ & 0.89$\times$ & 0.89$\times$  & 0.87$\times$ & 0.86$\times$ & 0.85$\times$ & 0.86$\times$ & 0.85$\times$ & 0.85$\times$ \\\midrule
         \multirow{2}{*}{{PGJ} (b=3)} & Iters & 1.07$\times$ & 1.09$\times$ & 1.09$\times$ & 1.09$\times$ & 1.10$\times$ & 1.07 $\times$& 1.07$\times$ & 1.08$\times$ & 1.08$\times$ & 1.11$\times$ \\ 
                                            & Time  & 1.01$\times$ & 1.05$\times$ & 1.05$\times$ & 1.07$\times$  & 1.04$\times$ & 1.02$\times$ & 1.02$\times$ & 1.03$\times$ & 1.03$\times$ & 1.05$\times$ \\\midrule
         \multirow{2}{*}{{HGJ} (b=3)} & Iters & 1.05$\times$ & 1.07$\times$ & 1.07$\times$ & 1.07$\times$ & 1.07$\times$ & 1.06$\times$ & 1.07$\times$ & 1.06$\times$ & 1.05$\times$ & 1.07$\times$ \\ 
                                       & Time  & 1.01$\times$ & 1.03$\times$ & 1.04$\times$ & 1.05$\times$ & 1.03$\times$ & 1.01$\times$ & 1.01$\times$ & 1.02$\times$ & 1.01$\times$ & 1.03$\times$ \\ \midrule
        \end{tabular}
        \caption{Comparison over different languages in terms of speedup and iterations on MBart50. Arrows indicate the direction of translation. Qualitative results and BLEU scores are available in the appendix \ref{sec:add_res_apex}.}
        \label{tab:mbart_lang}
        
\end{table*}
\subsection{Quality Guarantees}

Compared to NAT methods which do not have any quality guarantee since a novel parallel model is trained from scratch, our formulation guarantees to have the same quality of using autoregressive decoding with the same MT model. 
System \eqref{eq2} is known in literature as a \textit{triangular system} of $m$ equations with $m$ variables, this characterization allows to state an important property.
\begin{proposition}
\label{prop:1}
 Algorithms \ref{alg:pj}, \ref{alg:pgj}, \ref{alg:phgj} converge and yield the same results of greedy autoregressive decoding in at most $m$ parallel iterations, for any initialization and providing stopping condition \eqref{eq:stopping}.
\end{proposition}
We refer the reader to \citet{song:2021} for a formal proof. Intuitively, with $m$ steps the algorithm used the same number of iterations of autoregressive, hence the final solution is the same regardless the initialization. In this worst case, the wall-clock time is the same but in general the algorithm reach the stopping condition earlier with a lower wall-clock time and overall speedup.
\subsection{DDG\textit{viz}}
\label{sec:ddg_viz}
Equation \ref{eq:autoregressive} models the dependency between tokens in the decoding phase. In the classical autoregressive mode, each token depends on all the previous ones for the generation. However, it is possible to show that this dependency is actually relaxed (i.e., not all tokens depends on all the previous ones), thus it would be interesting to visualize the actual distribution $p_\theta\left(y_{i} \mid \cdot, \mathbf{x}\right)$ learned by an existing MT model. To this end, we build the Decoding Dependency Graph visualizer (DGG\textit{viz}) to visualize the dependency graph of tokens in the decoding phase. In the standard autoregressive decoding this graph is a fully-connected chain where the $i$-th token is connected to all the previous tokens, starting from the encoding $\mathbf{x}$: to decode $y_{i}$ you need to decode first $y_{1}, \ldots, y_{i-1}$. Instead we show that there are skipping connections between independent tokens that can be visualized with DGG\textit{viz}.
We detail DGG\textit{viz} with an example in section \ref{sec:decviz}.

\begin{table*}[t]
\scriptsize
\centering

\begin{tabular}{l|ccc|cc|ccc}
\multicolumn{1}{c}{\multirow{2}{*}{\textbf{Method}}} & \multicolumn{3}{c}{\textbf{Requirements}} & \multicolumn{2}{c}{\textbf{WMT14}} & \multicolumn{3}{c}{\textbf{Efficiency}} \\
\multicolumn{1}{c}{}   & Arch & Loss & seq-KD &{Speed} $\uparrow$ & {BLEU} $\uparrow$ & Train FLOPs  $\downarrow$ & Total FLOPs $\downarrow$ & FLOPs / Speed $\downarrow$ \\\hline

\textbf{Parallel Decoding - HGJ (Ours)} &\color{teal}{No}&\color{teal}{No}&\color{teal}{No}& 1.34$\times$ & \cellcolor{green!25} $28.24$& \textbf{0}& \textbf{2.53e+13} & \textbf{1.89e+13}\\
\hdashline
SUNDAE $^\dagger$\citep{Savinov:2022} & \color{red}{Yes}& \color{teal}{No}& \color{teal}{No} & 1.4$\times$ & \cellcolor{green!25} $28.46$ & 5.27e+21 & 5.27e+21 & 3.77e+21
\\
ShallowDec (12-1) \cite{kasai2021deep} & \color{red}{Yes}& \color{teal}{No} & \color{teal}{No} & 1.4$\times$ & $26.90$& 1.02e+19 & 1.02e+19 & 7.30e+18
\\
Semi-NAT \citep{wang-etal-2018-semi-nat} & \color{red}{Yes}& \color{teal}{No}& \color{red}{Yes} & 1.5$\times$ & $26.90$ & 1.55e+17 & 1.55e+17 & 1.03e+17
\\
DisCo \cite{Kasai:2020} & \color{red}{Yes} & \color{red}{Yes} & \color{red}{Yes, Big} & 3.5$\times$ & $27.34$& 4.06e+19 & 4.06e+19 & 1.16e+19
\\
DSLP \citep{Huang:2021} & \color{red}{Yes} & \color{red}{Yes} & \color{red}{Yes} & 14.8$\times$ & $27.02$& 1.93e+19 & 1.93e+19 & 1.31e+18
\\
F-VAE \citep{Gu:2021} & \color{red}{Yes} & \color{red}{Yes} & \color{red}{Yes, Big} & 16.5$\times$ & $27.49$& 4.06e+19 & 4.06e+19 & 2.46e+18
\\
\hline
\end{tabular}

\caption{Comparison of different methods for parallel MT on WMT14 En-De. Results are ordered by speed, highlighted in green the two highest BLEU scores, $^\dagger$ indicates diffusion models.
Existing methods require training, architecture modifications, additional losses to force parallel translation, and distillation from an additional MT transformer model ("Big" indicates the size).
Details on FLOPs computation are available in the Appendix \ref{sec:flops_apex}.
}
\label{tab:3}
\vspace{-0.3cm}
\end{table*}

\section{Experiments}

\subsection{Experimental Settings}

\paragraph{Datasets.}

We evaluate our approach using standard evaluation datasets proposed for parallel MT \cite{gu:2018}: WMT14 English-German [En-De], WMT16 English-Romanian [En-Ro] \cite{bojar-EtAl:2014:W14-33, bojar-EtAl:2016:WMT1}. Additionally, we tested our method on different language pairs with varying (low-medium) resources: IWSLT15 (English-Vietnamese [\text{En}-\text{Vi}]) \cite{tran-etal-2015-english}, IITB (English-Hindi [\text{En}-\text{Hi}]) \cite{kunchukuttan-etal-2018-iit}, WMT17 (English-Finnish [\text{En}-\text{Fi}]) \cite{bojar-EtAl:2017:WMT1}, FLORES-101 (English-Italian [\text{En}-\text{It}]; English-French [\text{En}-\text{Fr}]) \cite{flores101}. All the datasets are evaluated in both directions.
\paragraph{Evaluation.} All the evaluations are performed using the official test split for each dataset, downloaded using Huggingface dataset library  \cite{lhoest-etal-2021-datasets}. No training or hyperparameters tuning is performed.
We use SacreBLEU to evaluate the translation quality \cite{Papineni:2002, post-2018-call}.
We measure speedup in wall-clock time and iterations w.r.t. the same autoregressive model.
GPU times are calculated after calling \texttt{torch.cuda.synchronize()}.
All the experiments were performed by caching the past Keys and Values of the transformer to further speed up the computation \cite{DBLP:journals/corr/RamachandranPKB17} and in the online inference setting with batch size equal to $1$. For the Jacobi and GS-Jacobi algorithms, we assume to know beforehand the length $m$ of the target and measure the speedup in the ideal condition. For the Hybrid GS-Jacobi algorithm, we set $h$ equal to the maximum (i.e., the stopping condition is triggered within a parallel block) to decouple the effective speedup regardless of the length produced by the initialization function (see Section \ref{sect:paralledec}). We remark that  HGJ does not assume to know beforehand the target length and is applicable to real MT translation scenarios.

\paragraph{Model Configuration.}
We tested transformer models in the two standard configurations: base (512 model dimension, 6 attention layers for both encoder and decoder) and big (1024 model dimension, 12 attention layers for both encoder and decoder).
We used pretrained models of Opus \cite{TiedemannThottingal:EAMT2020} for the former and MBart50 \cite{mbart} for the latter.
Opus is a transformer base model (74M parameters) trained on language pairs from the homonymous dataset \cite{zhang2020improving}. MBart50 is a large multilingual transformer model fine-tuned for translation on 50 languages (610M parameters).
We tested the models on CPU since this is the default environment for MT models in production, except for the model MBart50 which runs on GPU.
We run the experiments on a standard 16-core machine, except for the scaling experiments. Additional specifications are available in Appendix \ref{sec:add_details}

\subsection{Algorithms Comparison}

In Table \ref{tab:1} we compare the proposed parallel decoding algorithms with the standard sequential autoregressive decoding baselines.
As we can observe, the fastest algorithms are PGJ Decoding (b=3) and HGJ Decoding (b=3) which are up to 34\% and 38\% times faster on Opus and up to 5\% and 8\% faster on MBart50, depending on the language pair. %
We note also that results empirically show that all the parallel decoding algorithms guarantee the same quality of greedy autoregressive decoding, as evidenced by the unchanged BLEU scores.
This is an experimental verification of the formal Proposition \ref{prop:1}.
The table also shows that the Beam Search algorithm with a beam size of 5 generally performs better in terms of BLEU score, although at a cost of speed. This difference in terms of BLEU is expected, as beam search is a heuristic search strategy, while our method is a decoding algorithm. We discussed better this aspect in the "Beam Search" paragraph. Nevertheless, beam search is  \(\sim\)30\% slower than greedy autoregressive and 63\% to 68\% slower than PGJ, depending on the model and language pair. This means that the proposed parallel algorithms allow trading a little translation quality (e.g., on en$\rightarrow$ro the difference between beam search and parallel decoding algorithms in BLEU is just $0.20$ points) for greater decoding speed.

Another aspect to note is that the algorithms PJ and PGJ (b=5) are sometimes slower than greedy autoregressive.
There are several factors that can influence the actual wall-clock time like how the underlying hardware schedule and execute the various operations, which might vary according to the architecture and the workload. In particular, longer sequences (e.g., the whole sentence in PJ or blocks of 5 tokens in PGJ) may require more memory to store, and the CPU/GPU may have to perform more memory accesses, which can slow down the computation (although theoretically it should happen in parallel). In the end, these computational overheads slow down the actual execution.
This is also the case for the difference in speedups between MBart50 and Opus.
We better investigated this aspect in the section "Computational Scaling" and report in the appendix results on a different architecture, with also results in terms of iterations speedups which are architecture agnostic.

\subsection{Analysis and Validation}
\label{sec:ana_valid}

\paragraph{Cross Languages.}In order to demonstrate the robustness of our decoding algorithms with respect to the translation languages, we leveraged the multilingual capabilities of the MBart50 model and selected a diverse range of language pairs for evaluation. The results, presented in Table \ref{tab:mbart_lang}, show that both PGJ and HGJ achieve a consistent speedup in comparison to the autoregressive decoding method, with an improvement ranging from 2-7\% for PGJ and 1-5\% for HGJ, regardless of the language pair used. Additionally, we observed a speedup in terms of iterations of 7-11\% for PGJ and 5-7\% for HGJ. These findings indicate that our algorithms have the potential to match or surpass the speedup in terms of wall-clock time by fully exploiting this saving in terms of iterations.
We note that, similar to the previous experiment, PJ suffers from an overhead problem. To the best of our knowledge, this is one of the first studies that have achieved a speedup in multilingual machine translation, concurrent with the work of \citet{song2022switchglat}, while this latter is significantly different in spirit and requirements (NAT model). We leave BLEU scores in the Appendix \ref{sec:add_res_apex} for space constraints together with qualitative results in different languages.
\paragraph{Computational Scaling.}

\begin{figure}[t]
    \centering
    \includegraphics[width=0.98\linewidth]{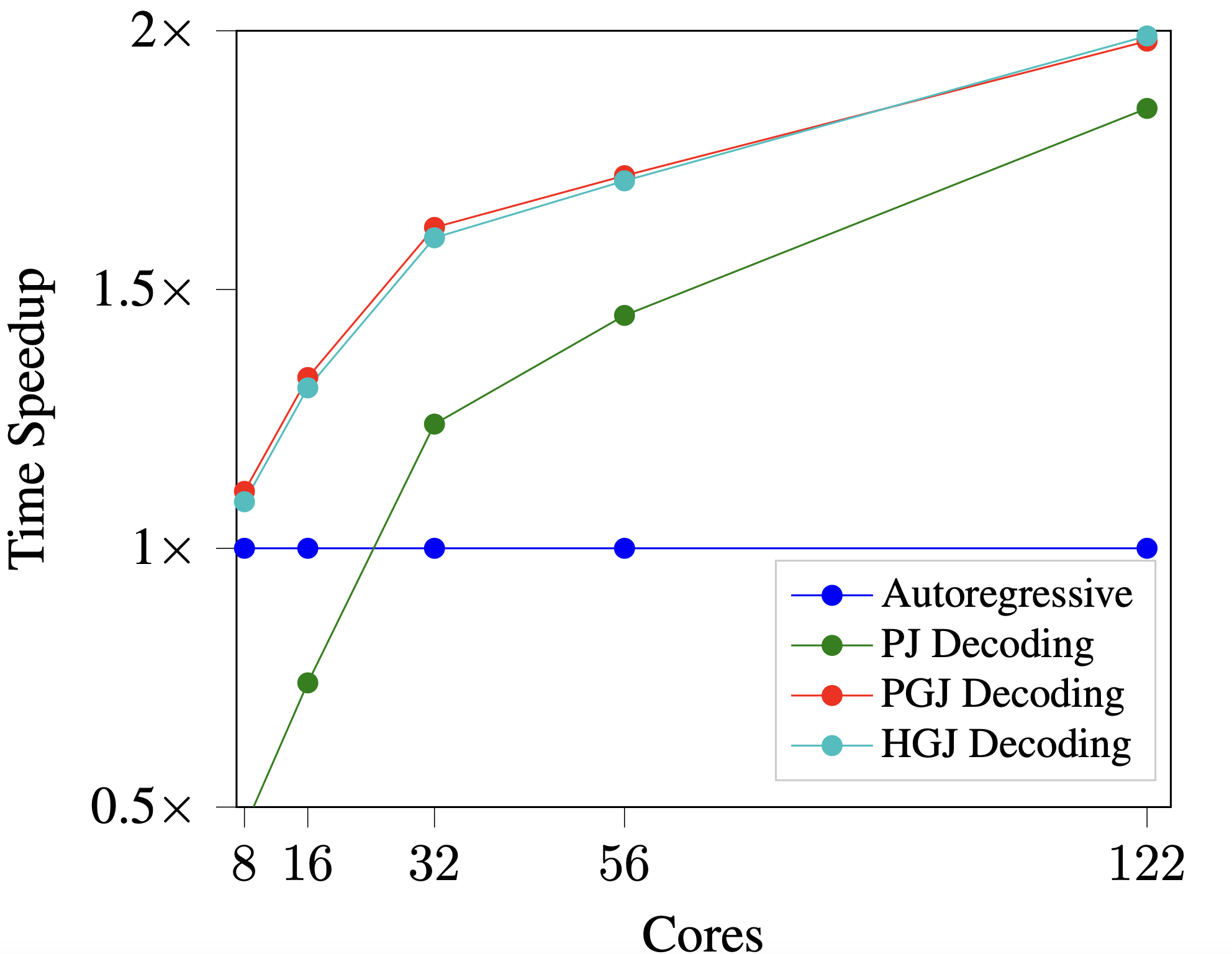}
    \caption{Scaling experiments on WMT16 En-De with PGJ and HGJ blocks = $3$.
Increasing the number of available resources (number of CPU cores) allows the methods to decrease the parallel overheads. As a result, the speedup increases and the methods scale.} 
\vspace{-0.5cm}
\label{fig:scaling}
\end{figure}

In Figure \ref{fig:scaling}, we present an analysis of the scalability of our proposed methods in relation to increasing computational resources. Starting with 8 cores, our methods demonstrate a slight improvement in terms of wall-clock time for PGJ and HGJ, with speedups of 1.11 and 1.09 respectively. On the other hand, this amount of resources is too restricting for PJ which needs to fit the whole sentence and thus achieve a score of $0.46$ due to the aforementioned overhead problem.
As the resources are increased, our method demonstrates the ability to effectively leverage hardware and significantly reduce decoding time, while the autoregressive baseline is constrained by sequential processing. 
With 122 cores, a substantial speedup of $1.98${$\times$} and $1.99${$\times$} is achieved for PGJ and HGJ respectively, while the autoregressive baseline is bounded by sequential processing at $1.00\times$.
It is important to note that this experiment does not simulate a real production system, but rather it is meant to show what results can be achieved when the underlying computation is properly optimized to run in parallel. In our case, we simulated this setting with increasing cores, nevertheless similar results can be achieved with 
additional software optimizations to further reduce latency and overheads \cite{Ahmed2022-da, kim-etal-2019-research} and increase the speed gain with parallel-optimized computations. 
Overall this experiment serves as a proof of concept for the capabilities of parallel decoding in contexts with limited overhead and shows a promising direction for further improvements.

\paragraph{Comparison with NATs.}
Table \ref{tab:3} reports the comparison of our parallel decoding algorithm with a selection of NAT methods for parallel MT. Following prior works, we report for each method the speedup relative to the autoregressive transformer base baseline  from their original paper \cite{xiao2022survey}.
It is worth noting that, although these methods can achieve higher speedups, they are very demanding in terms of computational resources which must be accounted for in a fair comparison. To estimate quantitatively this cost, we evaluated the number of floating point operations (FLOPs) required for training and inference on WMT14.

Results show that our method HGJ uses the least number of computational resources, even considering the additional cost at inference time. Relating the speedup obtained with the used resources (FLOPs/speed), our method still achieves the best cost-benefit ratio. Furthermore, NATs generally degrade the translation quality if compared to their autoregressive baseline. On the contrary, our method mathematically guarantees the same quality of autoregressive decoding, which is higher than standard NAT models. 

SUNDAE achieves BLEU of 28.46, but requires more resources than training RoBERTa \cite{liu2019roberta} on 16 TPUs (see Appendix \ref{sec:flops_apex}). Other methods require further elaborate techniques like profound architectural changes, additional losses to force parallel translation and sequence-level distillation from large autoregressive transformers \cite{Gu:2021}.
Our approach is a decoding method that does not involve any training or modification to the model and can be used to speed up existing models on standard desktop hardware.
\vspace{-0.05cm}

\paragraph{Speedup Analysis.}
We provide here a preliminary analysis of the factors responsible for the observed speedup in our method. We first distinguish between two types of speedup: wall-clock speedup and iterations speedup. The former is primarily driven by the parallelization capability of our method, as demonstrated in the "Computational Scaling" section. With parallel decoding, underlying operations can be optimized and fused to be executed fastly. Compared to \citet{sheng2023high}, our method allows parallelizing sequence operations ("row-by-row" setting).
The latter instead may vary consequently to several factors (e.g., model/vocabulary size, training data, language, etc). For this reason, we experimented with several variations of these factors (models Transformer Base vs. Big, vocabularies 58K Marian vs. 250K MBart50, languages, and hardware). While it is challenging to decouple different elements, our analysis point out several interesting insights. 
For example, we observed that iteration results on MBart50 are generally higher compared to Marian (Tables \ref{tab:mbart_lang}-\ref{tab:speed_iters}), possibly due to the finer-grained tokenization of MBart50. We also hypothesize that language and linguistic features, such as inflectionally rich or agglutinative/gendered languages, may influence iteration speedups.
To facilitate this type of analysis, we developed DDG\textit{viz}, which we believe will be useful for research in this area.

\begin{figure}[t]
\begin{center}
\includegraphics[width=0.9\columnwidth]{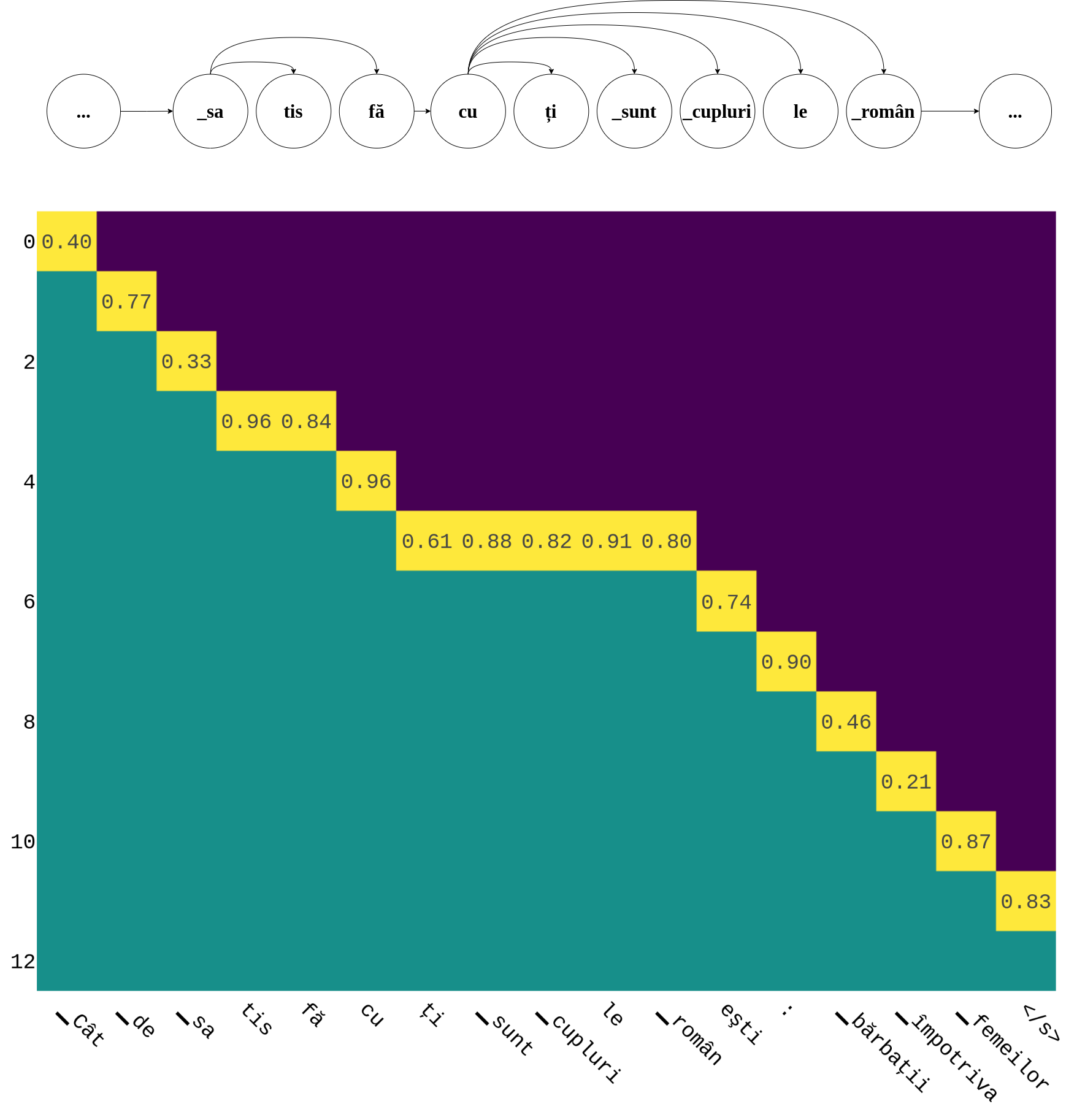}

\caption{\textbf{DDG\textit{viz}.} Visualization of the translation En-Ro: "How satisfied are the Romanian couples: men versus women"$\to$"Cât de sa\hl{tisfa}cu\hl{ti sunt cuplurile roman}ești: bărbații împotriva femeilor". (Highlighted tokens decoded in parallel).
\textbf{On top}: the Decoding Dependency Graph, omitting redundant edges on non-parallel tokens to ease visualization. \textbf{On bottom}: DDG\textit{viz} shows at each Parallel Jacobi iteration (vertical axis) which tokens have been generated in parallel (horizontal axis) with the corresponding probability (cell number).
}
\vspace{-0.7cm}
\label{fig:visualizer}
\end{center}

\end{figure}

\paragraph{Visualizing Parallel Decoding.}

\label{sec:decviz}

In previous experiments, we demonstrated that parallel decoding is feasible. This suggests that the dependency learned by the model between certain tokens is relaxed, as some tokens can be decoded in parallel. Analyzing and understanding when this happens allows shedding light on the behavior of existing models and a separate study focused on this issue would be needed. In this work, we lay the ground for a such study introducing the necessary inspection tools.
While we have already introduced DDG\textit{viz} in Section \ref{sec:ddg_viz}, in this experiment we show how it works and how it can be used with a practical example.
In summary, the DDG\textit{viz} visualizer allows to show the \textit{real} decoding distribution $p_\theta\left(y_{i} \mid \cdot, \mathbf{x}\right)$ learned by a MT model.
This decoding distribution is plotted as a graph, where a connection indicates the dependency $p_\theta(y_{i} \mid \cdot)$, by using Parallel Jacobi decoding.
At each PJ decoding iteration (vertical axis of Figure \ref{fig:visualizer}), DDG\textit{viz} keeps track of which tokens have been correctly decoded w.r.t. the gold autoregressive reference of the model, showing the tokens correctly decoded and the probability of each one (horizontal axis). 
Figure \ref{fig:visualizer} shows DDG\textit{viz} applied on an example.
The example shows that for $y_{4} = \_sa$ it is possible to decode more than one token in parallel $y_{5} = tis$, $y_{6} = fa$, hence here the decoding of $y_{6}$ does not depend on the decoding of $y_{5}$ - $p_\theta\left(y_{6} \mid \mathbf{y}_{1:\mathbf{4}}, \mathbf{x}\right)$.
We observed this phenomenon frequently, explaining the speedups in the previous experiments.  
The example also shows that the model is able to decode five tokens in parallel after $y_{7} = \_cu$.
This is a peculiar case since the model, given \textit{"How satisfi\_"}, is generating all at once \textit{"\_ed are the Romanian couples"} (proposed here in English for better readability, original version in Romanian is available in Figure).
This example indeed shows how DDG\textit{viz} can be used to highlight possible biases encoded in the model as it is not clear how the model can be so confident (see cell probability) that after "satisfied" the most straightforward tokens to decode are "Romanian couples" \cite{chang-etal-2019-bias, 10.1162/tacl_a_00401}.
We leave other use cases for future works and show
in Appendix \ref{sec:add_res_apex} several visualizations with equally interesting phenomena.

\section{Conclusions}
\label{sec:conclusion}

In this paper, we showed that is possible to speed up \textit{existing} machine translation models by simply changing the decoding procedure with a parallel formulation. We introduced three parallel decoding methods which achieve consistent speedups without requiring any training, modifications, or quality loss. Our solution is orthogonal to previous approaches proposed in literature which are demanding in terms of data, computational resources, and engineering effort.   
This makes it particularly useful in limited-resource scenarios when one or all of these requirements are not satisfiable and alternatives like NATs cannot be deployed.
While our method is not without shortcomings, it represents a valuable first step in the development of parallel decoding algorithms for machine translation that can be seamlessly integrated with any model.
We believe that further advancements in this area, including the exploration of optimal initialization procedures and stopping conditions, as well as the use of alternative parallel solvers for non-linear equations, will close the gap with learning-based techniques and continue to improve the efficiency and effectiveness of parallel decoding algorithms.

\section*{Acknowledgements}
We would like to thank Sébastien Bratières for his throughout feedback provided on this project. This work is supported by Translated with an Imminent Research Grant, ERC Starting Grant No. 802554 (SPECGEO), and PRIN 2020 project n.2020TA3K9N "LEGO.AI". Riccardo Marin is also supported by an Alexander von Humboldt Foundation Research Fellowship.

\section*{Limitations}
The proposed algorithms allow to speed up an existing model out-of-the-box, without any modification or retraining. However, there are some considerations to bear in mind when using parallel decoding in order to have a speedup in terms of wall-clock time. Firstly, as the name implies, the method executes the decoding phase in parallel. Therefore, to appreciate the speedup one should be able to run computations in parallel. Using parallel decoding without parallel resources or parallel-optimized software may increase wall-clock time due to overheads, leading to a waste of computation. This is further discussed in Section \ref{sec:ana_valid} "Computational Scaling".
The reported wall-clock time results are thus to be considered within the scope of the experimental setup proposed in this paper and they may vary depending on the underlying hardware and software.
Secondly, the method allows speedup of the decoding by scaling on parallel resources. This implies an additional computational cost during the inference phase to achieve a speedup. While using parallel decoding, one should consider a trade-off between the desired acceleration and the utilization of computational resources.
Thirdly, since our method performs the decoding in parallel, as for NAT systems, it is difficult to combine it with Beam Search. Beam Search is inherently a dynamic programming algorithm and it is not possible to efficiently maximize the joint probability of the large search space without using sequential intermediate computations. We better explain this aspect in the next paragraph.
\paragraph{Beam Search.}
Beam search is widely employed to enhance the translation quality in MT \cite{sutskever2014sequence,bahdanau2014neural} as well as in other domains such as audio \cite{reddy,postolache2023latent}. However, it is an inherently sequential procedure that stores partial joint probabilities of the entire sequence (beams) while progressing with autoregressive decoding. 
Determining the maximal joint probability of all sequences in parallel is a challenging task, equivalent to a full maximum a posteriori (MAP) estimation. This is an open research problem and it is also an issue for NAT methods. NAT methods patch up this limitation with sequence-level KD which has the advantage of "not requiring any beam search at test-time" \cite{kim2016sequence} thanks to learning and distillation from large models.
Since our method is a decoding algorithm, we cannot use the same approach without learning. Nevertheless, the quality guarantee allows our methods to have performance on par with greedy autoregressive and generally better than a NAT model. We think of our method, not as a replacement for beam search, but rather as a way to obtain a speedup at inference time that is a middle ground between autoregressive greedy decoding (high quality, no requirements, no speed) and NATs
(quality compromises, increasing requirements with increasing speed).
Future works might address the quality gap with beam search by combining parallel decoding with alternative techniques like Minimum Bayes Risk \cite{eikema-aziz-2020-map}.

\section*{Ethics Statement}
Increasing the inference speed of MT can positively impact society by giving people a fast and good translation. This will enable people from different language backgrounds to communicate with each other and help remove cultural and trade barriers. 
As demonstrated by comparing the number of FLOPs in Table 3, our method uses fewer resources compared to alternatives and thus has a smaller carbon footprint, making it a more sustainable choice \cite{strubell-etal-2019-energy}. Furthermore, since our method does not involve training procedures or change the quality of results, we do not introduce any societal bias (e.g. racism, sexism, homophobia) into the translations. The latter, however, can be introduced through data in the training of the backbone autoregressive models and NATs. It is the task of those who train these models to mitigate this problem. DDG\textit{viz} can also help investigate and visualize some potential harmful biases encoded in the model like in Figure \ref{fig:visualizer}.

\bibliography{acl2023}
\bibliographystyle{acl_natbib}
\appendix

\setlength{\tabcolsep}{4pt}
\section{Algorithms details}
We propose here the pseudocode of Algorithms \ref{alg:pgj} and \ref{alg:phgj} due to space limitations in the main body of the paper.

\begin{algorithm}[t]
\caption{Parallel GS-Jacobi Decoding\label{alg:pgj}}
\textbf{Input:} $\mathbf{x}=(x_1, \dots, x_n)$, $p_\theta$, ${b}$\\
\textbf{Output:} $\mathbf{y}=(y_1, \dots, y_m)$
\begin{algorithmic}[1]
\State $\mathbf{y} \gets \textsc{InitT}(\mathbf{x})$
\State $m \gets len(\mathbf{y})$
\State $i \gets 1$
\While{$i \leqslant m$}
\State $\mathbf{o} \gets copy(y_{i:i+b})$
\State ${\mathbf{y}_{i:i+b}} \gets \arg \max(p_\theta({\mathbf{y}_{i:i+b}}| \mathbf{y}_{1:i+b}, \mathbf{x}))$
\State $stop \gets \textsc{StopC}({o}, y_{i:i+b})$
\If{$stop$}
\State $i \gets i+b$
\State{break}
\EndIf
\EndWhile\\
\Return $\mathbf{y}$
\end{algorithmic}
\end{algorithm}

The function $copy(y_{i:i+b})$ creates a copy of the tensor in input detached from the source. This is done in practice to avoid the overwriting of pointers to the same memory location. Function $\textsc{CheckEOS}(y_{i:i+b})$ returns the index of the token EOS in the block if present, else $-1$. Function $\textsc{CheckEOS}(y_{i})$ returns $True$ if the tokes in exactly the token EOS, else $False$. The function $\arg \max$ selects from the model distribution over the vocabulary the index (token) with maximum probability. This procedure is done for all the tokens in parallel, in the case of parallel decoding, or for just a single token in the case of autoregressive decoding. Generally, the output is the prediction for the next token; hence it should be shifted left before the reassignment to a variable. We omitted this implementation detail for clarity.

\section{Additional implementation details}
\label{sec:add_details}
We run Opus experiments in table \ref{tab:1} on an AMD EPYC Milan with 16 cores at 2.45 GHz and 64GB of RAM (accessible on Google Cloud - \texttt{c2d-standard-16)}.
For the scalability experiment in figure \ref{fig:scaling}, we also used Google Cloud instances with an increasing number of cores (referred to as \texttt{c2d-standard-XX}, where \texttt{XX} is the number of used cores).
Experiments with MBart50 on table \ref{tab:1}, \ref{tab:mbart_lang} and \ref{tab:speed_iters} are performed on a Desktop machine with Ubuntu 20.04.4 LTS, AMD Ryzen 9 3900X 12-Core Processor, 32GB of RAM, and a Palit Nvidia 3090 GPU.
Additional experiments with Opus in table \ref{tab:speed_iters} are also performed on this machine.
Models are implemented in Pytorch 1.11.0
\cite{NEURIPS2019_9015} and the Huggingface Transformer library \cite{wolf-etal-2020-transformers}.
We used python 3.8 and NVIDIA-SMI Drivers 510.73.05 with CUDA version 11.6.
For OPUS we used Huggingface models available on the hub under the tag \texttt{Helsinki-NLP/opus-mt-\{src\}-\{tgt\}} except for the language pair Ro-En where we used the model \texttt{Helsinki-NLP/opus-mt-roa-en} and the pair En-De where we used the checkpoint \texttt{opus-2021-02-22} \footnote{https://object.pouta.csc.fi/Tatoeba-MT-models/eng-deu/opus-2021-02-22.zip}.
For the model MBart50, we used the \texttt{facebook} pre-trained model available on the hub with the tag \texttt{mbart-large-50-many-to-many-mmt}. Since this is a multilingual model, we prepend the source and target language tag corresponding properly to the language pair to be translated.
We report results for a single run over the test dataset since we found low variance in estimates with multiple runs which can be calculated by simply varying the corresponding parameter in the \texttt{config.yaml} file.
For each dataset, we used the official test split via the Huggingface dataset library  \cite{lhoest-etal-2021-datasets}. Datasets statistics are reported in table \ref{tab:dataset_stastistics}.

\begin{algorithm}[t]
\caption{Hybrid GS-Jacobi Decoding \label{alg:phgj}}
\textbf{Input:} $\mathbf{x}=(x_1, \dots, x_n)$, $p_\theta$, ${b}$\\
\textbf{Output:} $\mathbf{y}=(y_1, \dots, y_m)$
\begin{algorithmic}[1]
\State $\mathbf{y} \gets \textsc{InitT}(\mathbf{x})$
\State $h \gets len(\mathbf{y})$
\State $i \gets 1$
\State $eos\_cond \gets False$
\While{$i \leqslant h$}
\State $\mathbf{o} \gets copy(\mathbf{y}_{i:i+b})$
\State ${\mathbf{y}_{i:i+b}} \gets \arg \max(p_\theta({\mathbf{y}_{i:i+b}}|\mathbf{y}_{1:i+b}, \mathbf{x}))$
\State $stop \gets \textsc{StopC}(\mathbf{o}, \mathbf{y}_{i:i+b})$
\State $eos\_ind \gets \textsc{CheckEOS}(\mathbf{y}_{i:i+b})$
\If{$stop$ \textbf{and} $eos\_ind > -1$}
\State $\mathbf{y} \gets \mathbf{y}_{1:eos\_ind}$
\State $eos\_cond \gets True$
\State{break}
\EndIf
\If{$stop$}
\State $i \gets i+b$
\State{break}
\EndIf
\EndWhile
\While{$eos\_cond \text{ } != True$}
\State ${y_{i}} \gets \arg \max(p_\theta({y_{i}}| y_{i-1}, \mathbf{x}))$
\State $i \gets i+1$
\State $eos\_cond \gets \textsc{IsEOS}(y_{i})$
\EndWhile\\
\Return $\mathbf{y}$
\end{algorithmic}
\end{algorithm}

\begin{table}[t]
\small
\centering

\begin{tabular}{l|c}
Dataset & \# Test \\ \hline
WMT 14 De-En \cite{bojar-EtAl:2014:W14-33}& 3003 \\
WMT 16 Ro-En \cite{bojar-EtAl:2016:WMT1}& 1999\\
WMT 17 Fi-En \cite{bojar-EtAl:2017:WMT1}& 3002\\
IWSLT 15 En-Vi \cite{tran-etal-2015-english} & 1046\\
IITB En-Hi \cite{kunchukuttan-etal-2018-iit} & 2507\\
FLORES-101 En-It \cite{flores101} & 1012 \\
FLORES-101 En-Fr \cite{flores101}& 1012\\

\hline
\end{tabular}

\caption{Data Statistic}
\label{tab:dataset_stastistics}

\end{table}

\section{FLOPs calculation details}

\label{sec:flops_apex}
We measured computational complexity using floating point operations (FLOPs), which, as the name imply, counts the number of floating point operation performed by a model. This is a standard metric used in literature to measure hardware-agnostic complexity. This means that hardware and software optimizations are not counted in the score \cite{wu2016google,kim-etal-2019-research}. We used the ELECTRA flops calculator\footnote{https://github.com/google-research/electra/blob/master/flops\_computation.py} inserting the number of parameters and the number of training step performed for each model analyzed in table \ref{tab:3} according to the training specification in each paper.
For inference FLOPs, we computed the decoding cost of each sentence in the testset of WMT14 En-De for each model.
For a scale reference, we report in here Table \ref{tab:flops_add} training flops of other well-known architecture. The code package contains the scripts to replicate all the experiments.

\begin{table}[h!]
\small
\centering

\begin{tabular}{c|c|c|c}
Model & Train FLOPs & Infer. FLOPs & Total FLOPs\\\hline
Semi-NAT&1.55e17 & 2.08e13 & 1.55e17  \\
Shallow Dec. & 1.02e19 & 1.15e13 & 1.02e19  \\
DSLP& 1.93e19 & 1.58e13 & 1.93e19 \\ 
F-VAE & 4.06e19 & 1.58e13 & 4.06e19  \\
DisCo & 4.06e19 & 1.58e13 & 4.06e19 \\
SUNDAE&5.27e21 & 1.58e14 & 5.27e21  \\
\hdashline
BERT base & 6.43e19 & - & - \\
BERT large & 1.92e20 & - & - \\
RoBERTa & 3.19e21 & - & - \\
\hline
\end{tabular}

\caption{FLOPs comparison with other models.}
\label{tab:flops_add}

\end{table}

\section{Additional results}
\label{sec:add_res_apex}
\begin{table*}[t]
\centering
\small
\begin{tabular}{l|cc|cc|cc|cc}
\multirow{2}{*}{\textbf{Decoding Algorithm}}&
\multicolumn{2}{c}{\textbf{en$\rightarrow$de}} &  
\multicolumn{2}{c|}{\textbf{de$\rightarrow$en}}  & 
\multicolumn{2}{c}{\textbf{en$\rightarrow$ro}} &  
\multicolumn{2}{c}{\textbf{ro$\rightarrow$en}} \\
 &  Time\ & Iters& Time & Iters & Time & Iters & Time & Iters \\
\hline
\textbf{Opus} & & & & & & & &\\
Greedy Autoregressive & $1.00\times$ & $1.00\times$&$1.00\times$& $1.00\times$ & $1.00\times$ & $1.00\times$ & $1.00\times$ &  $1.00\times$\\
Beam Search (beam = 5) & $0.71\times$ & $1.00\times$&$0.71\times$& $1.00\times$ & $0.70\times$ & $1.00\times$ & $0.72\times$ &  $1.00\times$\\
\rowcolor[gray]{.90} PJ Decoding & $0.72\times$& $1.03\times$ & $0.74\times$& $1.04\times$ & $0.69\times$ & $1.04\times$& $0.67\times$& $1.03\times$ \\
\rowcolor[gray]{.90} PGJ Decoding (b = 3) & $\mathbf{1.16\times}$& $1.04\times$ &$\mathbf{1.19\times}$& $1.07\times$ & ${1.17\times}$ & $1.05\times$  & $\mathbf{1.17\times}$ & $1.03\times$ \\
\rowcolor[gray]{.90} HGJ Decoding (b = 3) & $\mathbf{1.16\times}$ & $1.04\times$& $\mathbf{1.19\times}$ & $1.06\times$ &$1.17\times$ & $1.05\times$ &$\mathbf{1.17\times}$ & $1.03\times$ \\ 
\hline
\textbf{MBart50} & & & & & & & &\\
 Greedy Autoregressive & $1.00\times$ & $1.00\times$&$1.00\times$& $1.00\times$ & $1.00\times$ & $1.00\times$ & $1.00\times$ &  $1.00\times$\\
Beam Search (beam = 5) & $0.76\times$ & $1.00\times$&$0.77\times$& $1.00\times$ & $0.77\times$ & $1.00\times$ & $0.76\times$ &  $1.00\times$\\
\rowcolor[gray]{.90} PJ Decoding & $0.88\times$& $1.03\times$ & $0.88\times$& $1.03\times$ & $0.86\times$ & $1.04\times$ & $0.85\times$& $1.03\times$ \\
\rowcolor[gray]{.90} PGJ Decoding (b = 3) & $\mathbf{1.06\times}$& $1.10\times$ &$\mathbf{1.08\times}$& $1.11\times$ & $\mathbf{1.03\times}$ & $1.08\times$ & $\mathbf{1.04\times}$ & $1.11\times$ \\
\rowcolor[gray]{.90} HGJ Decoding (b = 3) & $1.05\times$ & $1.07\times$ & $1.07\times$ & $1.01\times$ &$1.01\times$ & $1.02\times$ &$1.02\times$ & $1.08\times$ \\ 
\hline
\end{tabular}
\caption{Comparison of parallel decoding algorithms (highlighted in grey) with sequential decoding using Opus (CPU) and MBart50 (GPU) on WMT14 and WMT16. Speed is showed here both in Time and Iterations w.r.t. the greedy autoregressive baseline.}
\label{tab:speed_iters}
\end{table*}

\begin{table*}[t]
    \small
    \centering
    
        \begin{tabular}{lllllllllll}
        
        \multicolumn{1}{c}{} & \multicolumn{2}{c}{\textbf{WMT17}} & \multicolumn{2}{c}{\textbf{IITB}} & \multicolumn{2}{c}{\textbf{IWSLT15}}& \multicolumn{4}{c}{\textbf{FLORES}}\\
        \multicolumn{1}{c}{} & \multicolumn{2}{c}{\textbf{En-Fi}} & \multicolumn{2}{c}{\textbf{En-Hi}} & \multicolumn{2}{c}{\textbf{En-Vi}} &  \multicolumn{2}{c}{\textbf{En-It}} & \multicolumn{2}{c}{\textbf{En-Fr}} \\
        \cmidrule(lr){2-3} \cmidrule(lr){4-5} \cmidrule(lr){6-7} \cmidrule(lr){8-9} \cmidrule(lr){10-11}
        \textbf{Dec. Algorithm} &  $\text{ }\leftarrow$ & $ \text{ }\rightarrow$ & $ \text{ }\leftarrow$ & $ \text{ }\rightarrow$ & $ \text{ }\leftarrow$ & $ \text{ }\rightarrow$ & $ \text{ }\leftarrow$ & $ \text{ }\rightarrow$ & $ \text{ }\leftarrow$ & $ \text{ }\rightarrow$ \\ \midrule
        Autoregressive  & $17.55$ & $25.34$ & $16.50$ & $24.70$ & $31.92$ & $33.94$ & $22.78$ & $26.38$ & $39.51$ & $38.90$ \\
        Beam Search  & $18.39$ & $26.04$ & $16.87$ & $25.24$ & $32.14$ & $34.59$ & $23.52$ & $26.80$ & $39.59$ & $39.21$ \\
        PJ & $17.54$ & $25.35$ & $16.50$ & $24.69$ & $31.92$ & $33.94$ & $22.78$ & $26.38$ & $39.50$ & $38.90$ \\
        PGJ (b=3) & $17.55$ & $25.35$ & $16.50$ & $24.70$ & $31.93$ & $33.94$ & $22.78$ & $26.38$ & $39.51$ & $38.90$ \\
        HGJ (b=3) & $17.55$ & $25.35$ & $16.50$ & $24.70$ & $31.93$ & $33.94$ & $22.78$ & $26.38$ & $39.51$ & $38.90$\\ \bottomrule
        \end{tabular}
        \caption{BLEU scores on MBart50.
        }
        \label{tab:mbart_lang_bleu}
\end{table*}

\begin{figure*}
    \centering
    \includegraphics[scale=0.4]{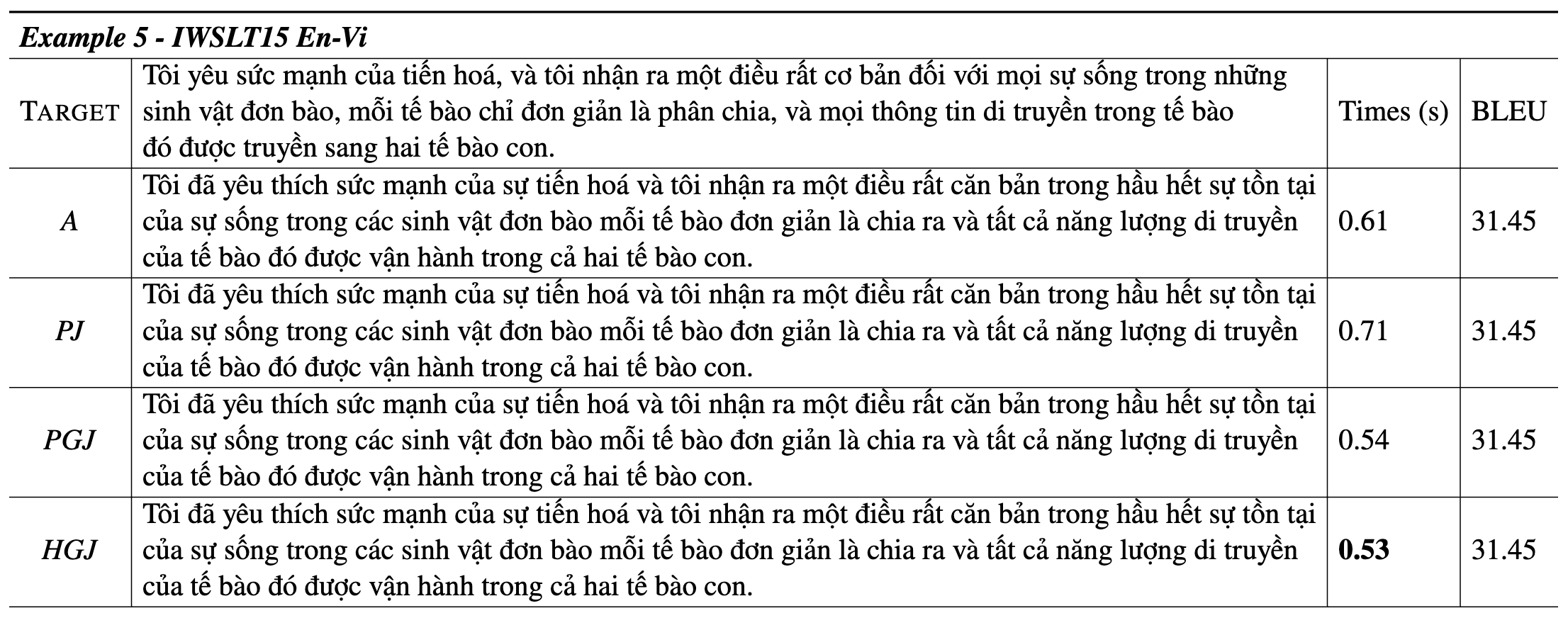}
    \captionsetup{labelformat=empty}
    \caption{Table 7: Translation examples generated with the autoregressive (A) and the different decoding algorithms proposed (PJ, PGJ, HGJ) on Opus (WMT datasets) and MBart50. The decoding time is shown in seconds.}
    \label{tab:qualres}
\end{figure*}

\begin{figure*}
\begin{subfigure}{.5\textwidth}
  \centering
  \captionsetup{width=.9\linewidth}
  \includegraphics[width=.8\linewidth]{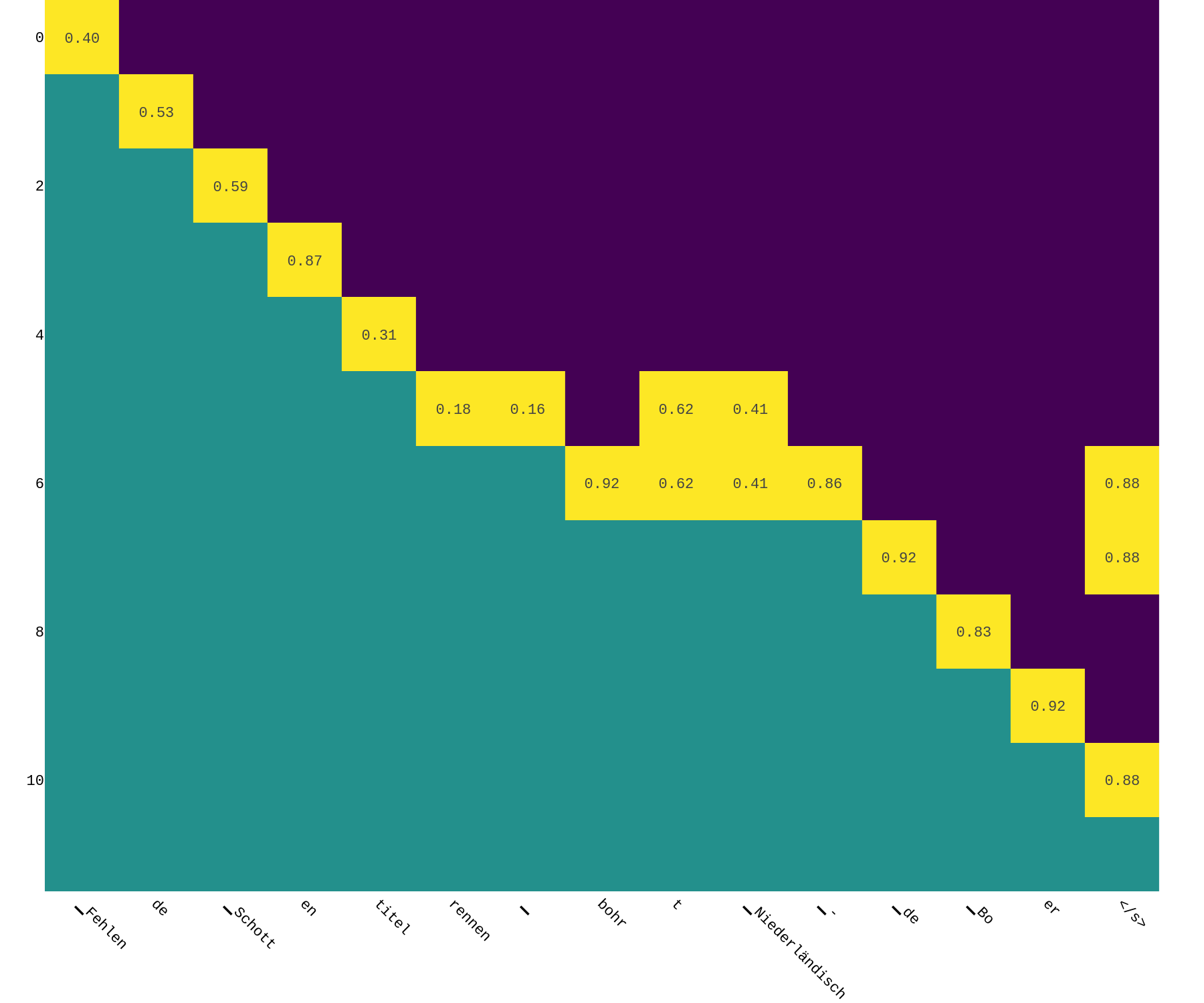}
  \caption{En-De: "Lack of Scots title race bores Dutch - de Boer"$\to$"Fehlende Schottentitel\hl{rennen }\hl{bohrt Niederlandisch -} de Boer"}
\end{subfigure}%
\begin{subfigure}{.5\textwidth}
  \centering
  \captionsetup{width=.9\linewidth}
  \includegraphics[width=.8\linewidth]{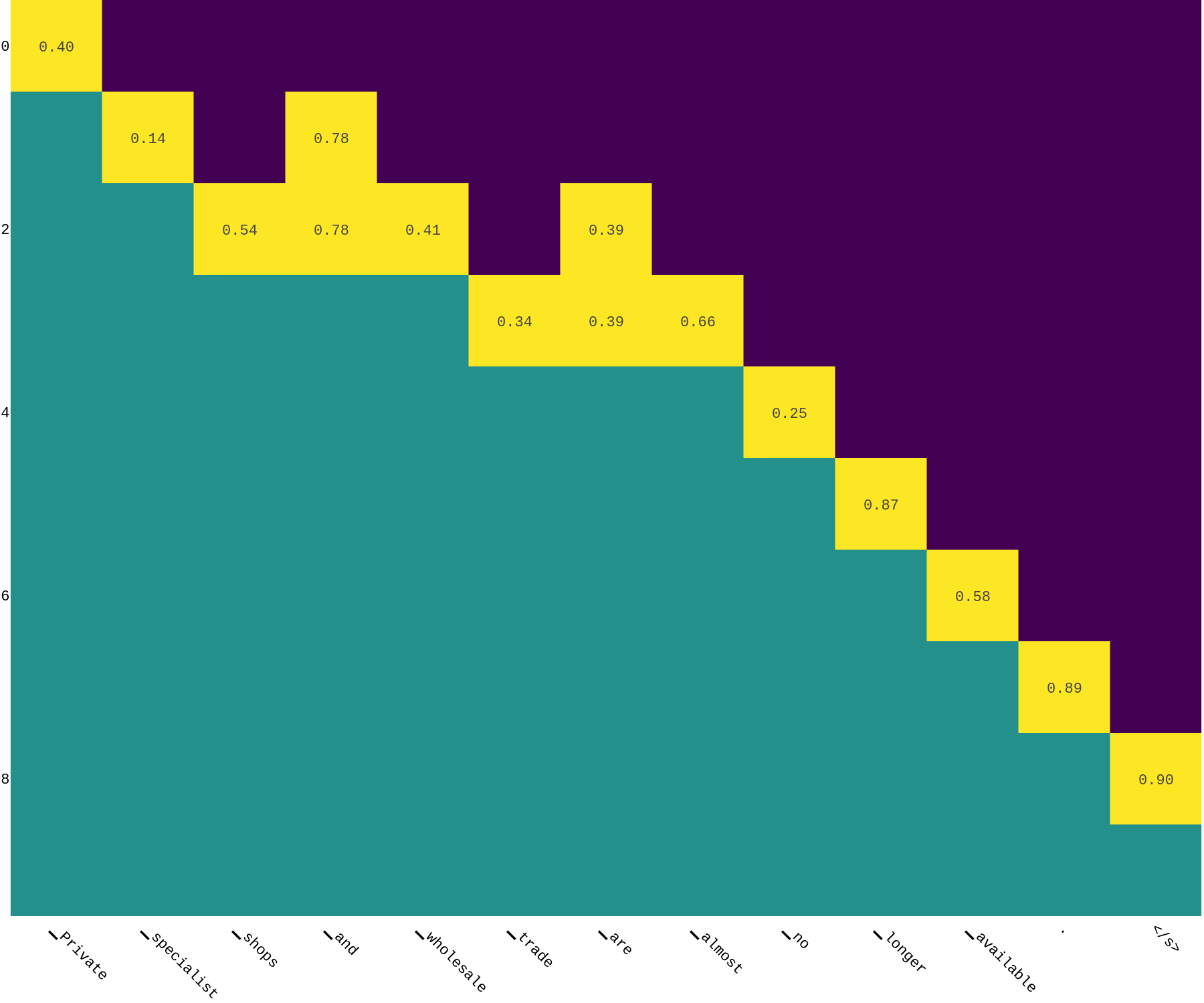}
  \caption{De-En: "Private Fachgeschafte und auch den Großhandel gibt es fast nicht mehr."$\to$"Private specialist \hl{shops and wholesale} \hl{trade are almost} no longer available."}
\end{subfigure}
\begin{subfigure}{.5\textwidth}
  \vspace{10pt}
  \centering
  \captionsetup{width=.9\linewidth}
  \includegraphics[width=.9\linewidth]{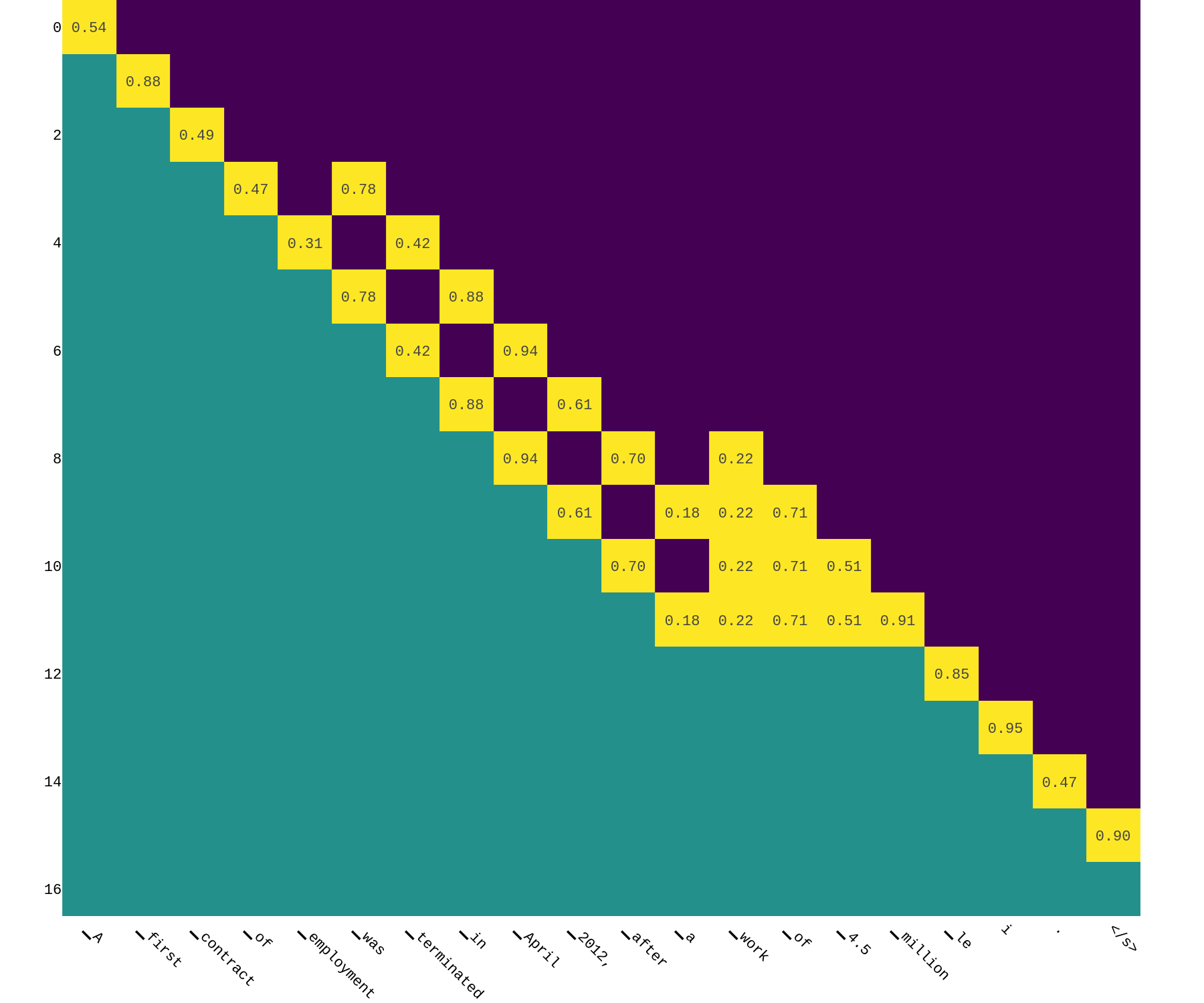}
  \caption{Ro-En: "Un prim contract de lucrări a fost reziliat în aprilie 2012, după ce se efectuaseră lucrări de 4,5 milioane lei."$\to$
  "A first contract of employment was terminated in April 2012, after \hl{a work of 4.5 million} lei."}
\end{subfigure}%
\begin{subfigure}{.5\textwidth}
  \centering
  \captionsetup{width=.9\linewidth}
  \includegraphics[width=.9\linewidth]{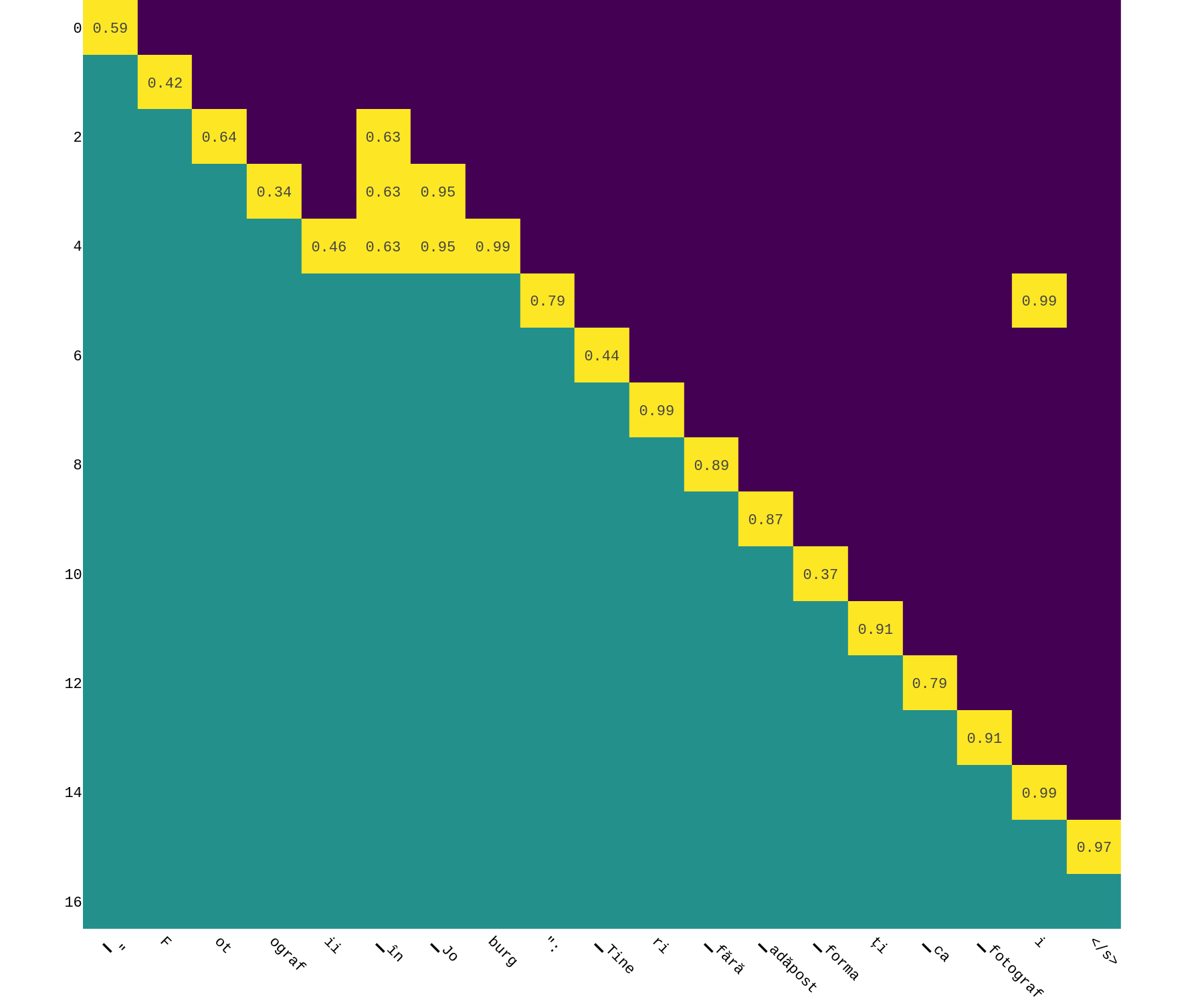}
  \caption{En-Ro: "`Shot in Joburg': Homeless youth trained as photographers"$\to$
  "``Fotograf\hl{ii in Joburg}'': Tineri fără adăpost formaţi ca fotografi"}
\end{subfigure}
\begin{subfigure}{.5\textwidth}
  \vspace{10pt}
  \centering
  \captionsetup{width=.9\linewidth}
  \includegraphics[width=.9\linewidth]{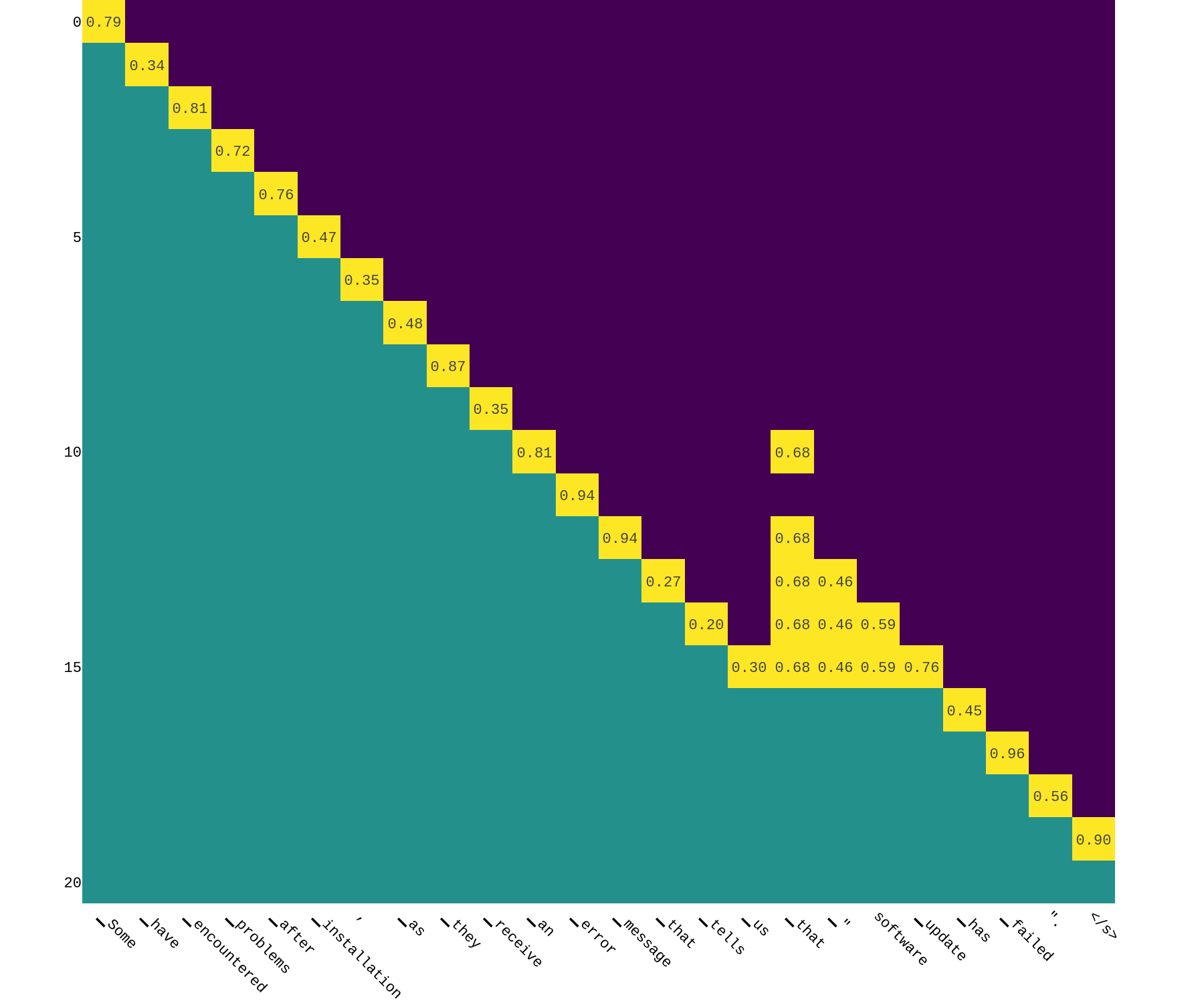}
  \caption{De-En: "Einige sind nach der Installation auf Probleme gestoßen, da sie eine Fehlermeldung erhalten, die mitteilt, dass die ``Software-Aktualisierung fehlgeschlagen'' ist."$\to$"Some have encountered problems after installation, as they receive an error message that tells \hl{us that ``software update} has failed''."}
\end{subfigure}%
\begin{subfigure}{.5\textwidth}
  \centering
  \captionsetup{width=.9\linewidth}
  \includegraphics[width=.9\linewidth]{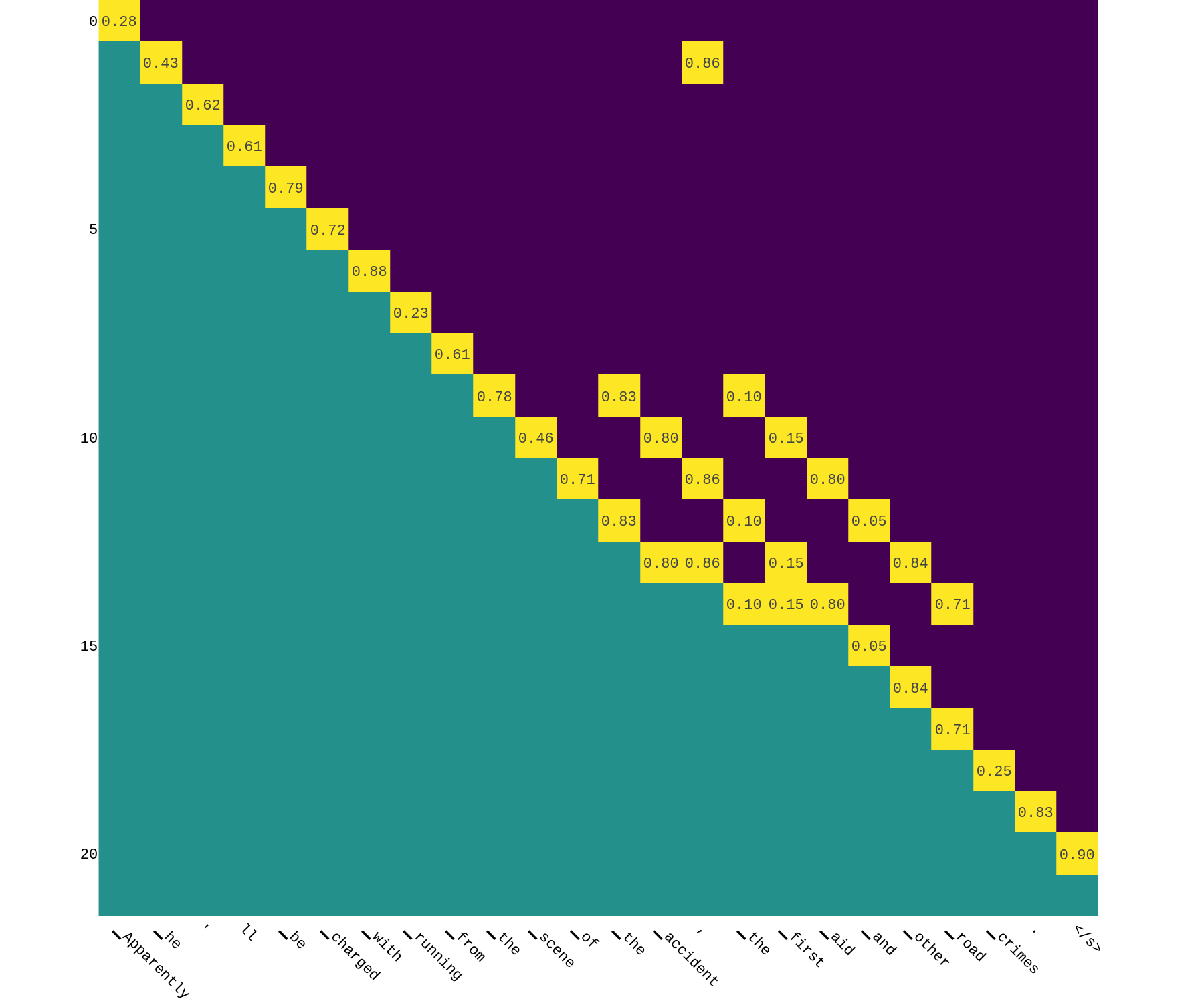}
  \caption{Ro-En: "Se pare că va fi acuzat de fugă de la locul accidentului, neoferirea primului ajutor și alte infracțiuni rutiere."$\to$
  "Apparently he'll be charged with running from the scene of the \hl{accident,} \hl{the first aid} and other road crimes."}
\end{subfigure}
\caption{DGG\textit{viz} additional visualizations}
\label{fig:ddg_additional}
\end{figure*}

We propose here additional results to the experiments in the paper that were omitted due to limitations constraints. %
Table \ref{tab:speed_iters} shows the same experiments of Table \ref{tab:1} in the main paper, proposed here on a standard desktop CPU with also the speedup in terms of iterations. It is possible to observe that in the case of MBart50 and PGJ there is a speedup of $8-11\%$ in terms of iterations compare to a time speedup of $3-8\%$. This means that there is room for improvement for our algorithm. 
Furthermore, results show that the time speedups are consistent also with standard desktop hardware.
Table \ref{tab:mbart_lang_bleu} shows the BLEU scores for the cross-lingual experiment. It is possible to observe that parallel decoding algorithms guarantee quality compared to greedy autoregressive and are not so distant from beam search. 
We show also here in table \ref{tab:qualres} some qualitative results for the experiments in table \ref{tab:mbart_lang}.
Finally, we propose additional visualizations using DGG\textit{viz} in Figure \ref{fig:ddg_additional}.

\newpage

\begin{table*}[h]
\centering
\resizebox{\textwidth}{!}{
\begin{tabular}{c|l|l|l}
\toprule
\multicolumn{4}{l}{\textbf{\textit{Example 1 - Wmt16 En-Ro}}}\\\hline
\textsc{Target} & \makecell[l]{Dl Corbyn va adresa primele dintre cele șase întrebări la care are dreptul la scurt timp după prânz; prestația\\ sa va fi probabil analizată îndeaproape de mass-media și parlamentarii laburiști.} &   Times (s) & BLEU\\\hline
\textit{A} & \makecell[l]{Dl Corbyn va ridica pentru a adresa prima dintre cele şase întrebări alocate la scurt timp după miezul zilei, iar\\ performanţa sa va fi probabil examinată îndeaproape de presă şi de parlamentarii laburişti.} & 0.51 & 19.71  \\ \hline
\textit{PJ} &  \makecell[l]{Dl Corbyn va ridica pentru a adresa prima dintre cele şase întrebări alocate la scurt timp după miezul zilei, iar\\ performanţa sa va fi probabil examinată îndeaproape de presă şi de parlamentarii laburişti.} &  0.56 & 19.71 \\ \hline
\textit{PGJ} & \makecell[l]{Dl Corbyn va ridica pentru a adresa prima dintre cele şase întrebări alocate la scurt timp după miezul zilei, iar\\ performanţa sa va fi probabil examinată îndeaproape de presă şi de parlamentarii laburişti.} & 0.45 & 19.71\\ \hline
\textit{HGJ} & \makecell[l]{Dl Corbyn va ridica pentru a adresa prima dintre cele şase întrebări alocate la scurt timp după miezul zilei, iar\\ performanţa sa va fi probabil examinată îndeaproape de presă şi de parlamentarii laburişti.} & \textbf{0.44} & 19.71 \\ \hline

\end{tabular}
}
\end{table*}

\begin{table*}[h]
\centering
\resizebox{\textwidth}{!}{
\begin{tabular}{c|l|l|l}
\toprule
\multicolumn{4}{l}{\textbf{\textit{Example 2 - Flores En-It}}}\\\hline
\textsc{Target} & \makecell[l]{Quando un piccolo gruppo di esseri viventi (una piccola popolazione) si separa dalla popolazione principale \\ alla quale appartiene (per esempio se si sposta oltre una catena montuosa o un fiume, o si sposta su una nuova \\ isola, rendendo quindi difficile un eventuale ritorno), esso si ritroverà probabilmente in un ambiente diverso da \\ quello in cui si trovava prima.} &   Times (s) & BLEU\\\hline
\textit{A} & \makecell[l]{Quando un piccolo gruppo di esseri viventi si separa dalla popolazione principale da cui provengono, come se \\ si muovano su una catena di montagne o su un fiume o se si trasferiscono su una nuova isola per non poter tornare \\ facilmente, si troveranno spesso in un ambiente diverso da quello in cui erano prima.} & 0.61 & 31.69  \\ \hline
\textit{PJ} &  \makecell[l]{Quando un piccolo gruppo di esseri viventi si separa dalla popolazione principale da cui provengono, come se \\ si muovano su una catena di montagne o su un fiume o se si trasferiscono su una nuova isola per non poter tornare \\ facilmente, si troveranno spesso in un ambiente diverso da quello in cui erano prima.} &  0.73 & 31.69 \\ \hline
\textit{PGJ} & \makecell[l]{Quando un piccolo gruppo di esseri viventi si separa dalla popolazione principale da cui provengono, come se \\ si muovano su una catena di montagne o su un fiume o se si trasferiscono su una nuova isola per non poter tornare \\ facilmente, si troveranno spesso in un ambiente diverso da quello in cui erano prima.} & \textbf{0.58} & 31.69\\ \hline

\textit{HGJ} & \makecell[l]{Quando un piccolo gruppo di esseri viventi si separa dalla popolazione principale da cui provengono, come se \\ si muovano su una catena di montagne o su un fiume o se si trasferiscono su una nuova isola per non poter tornare \\ facilmente, si troveranno spesso in un ambiente diverso da quello in cui erano prima.} &  0.59 & 31.69 \\ \hline
\end{tabular}
}
\end{table*}

\begin{table*}[h]
\centering
\resizebox{\textwidth}{!}{
\begin{tabular}{c|l|l|l}
\toprule
\multicolumn{4}{l}{\textbf{\textit{Example 3 - Wmt14 En-De}}}\\\hline
\textsc{Target} & \makecell[l]{Bei der diesjährigen Veranstaltung gibt es Auftritte von Wanda Sykes, Kathy Griffin und Bill Maher sowie auch \\von „Stand Up for Heroes“, einer jährlichen Musik- und Comedy-Benefizveranstaltung für Armeeveteranen im\\ Madison Square Garden, bei der unter anderem Bruce Springsteen, Jon Stewart, Roger Waters und Bill Cosby auftreten.} &   Times (s) & BLEU\\\hline
\textit{A} & \makecell[l]{Zu den diesjährigen Veranstaltungen gehören Auftritte von Wanda Sykes, Kathy Griffin und Bill Maher sowie\\ "Stand Up for Heroes", ein jährlicher Musik- und Komödie-Vorteil für Militärveteranen, im Madison Square Garden, mit\\ u.a. Bruce Springsteen, Jon Stewart, Roger Waters und Bill Cosby.} & 1.30 & 47.04 \\ \hline
\textit{PJ} &  \makecell[l]{Zu den diesjährigen Veranstaltungen gehören Auftritte von Wanda Sykes, Kathy Griffin und Bill Maher sowie\\ "Stand Up for Heroes", ein jährlicher Musik- und Komödie-Vorteil für Militärveteranen, im Madison Square Garden, mit\\ u.a. Bruce Springsteen, Jon Stewart, Roger Waters und Bill Cosby.} &  2.43 & 47.04 \\ \hline

\textit{PGJ} & \makecell[l]{Zu den diesjährigen Veranstaltungen gehören Auftritte von Wanda Sykes, Kathy Griffin und Bill Maher sowie\\ "Stand Up for Heroes", ein jährlicher Musik- und Komödie-Vorteil für Militärveteranen, im Madison Square Garden, mit\\ u.a. Bruce Springsteen, Jon Stewart, Roger Waters und Bill Cosby.} & 1.09 & 47.04\\ \hline
\textit{HGJ} & \makecell[l]{Zu den diesjährigen Veranstaltungen gehören Auftritte von Wanda Sykes, Kathy Griffin und Bill Maher sowie\\ "Stand Up for Heroes", ein jährlicher Musik- und Komödie-Vorteil für Militärveteranen, im Madison Square Garden, mit\\ u.a. Bruce Springsteen, Jon Stewart, Roger Waters und Bill Cosby.} &   \textbf{1.08} & 47.04\\ \hline
\end{tabular}
}
\end{table*}

\begin{table*}[h]
\centering
\resizebox{\textwidth}{!}{
\begin{tabular}{c|l|l|l}
\toprule
\multicolumn{4}{l}{\textbf{\textit{Example 4 - Flores En-Fr}}}\\\hline
\textsc{Target} & \makecell[l]{Cinq minutes après le début de l'exposition, un vent se met à souffler pour atteindre, environ une minute \\ plus tard, la vitesse de 70km/h... puis la pluie arrive, mais si forte et si grosse qu'elle frappe votre peau \\ comme une aiguille, puis la grêle tombe du ciel, les gens paniquent, crient et se roulent dessus.} &   Times (s) & BLEU \\\hline
\textit{A} & \makecell[l]{Cinq minutes après l'exposition, le vent commence à tourner, environ un minute plus tard, le vent atteint \\ 70 km/h, puis la pluie arrive, mais si forte et si grande qu'elle vous frappe la peau comme une aiguille, puis \\ le hail tombe du ciel, les gens paniquent, s'expriment et se courent l'un sur l'autre.} & 0.82 & 39.90  \\ \hline
\textit{PJ} &  \makecell[l]{Cinq minutes après l'exposition, le vent commence à tourner, environ un minute plus tard, le vent atteint \\ 70 km/h, puis la pluie arrive, mais si forte et si grande qu'elle vous frappe la peau comme une aiguille, puis \\ le hail tombe du ciel, les gens paniquent, s'expriment et se courent l'un sur l'autre.} &  0.94 & 39.90 \\ \hline
\textit{PGJ} & \makecell[l]{Cinq minutes après l'exposition, le vent commence à tourner, environ un minute plus tard, le vent atteint \\ 70 km/h, puis la pluie arrive, mais si forte et si grande qu'elle vous frappe la peau comme une aiguille, puis \\ le hail tombe du ciel, les gens paniquent, s'expriment et se courent l'un sur l'autre.} & 0.73 & 39.90\\ \hline
\textit{HGJ} & \makecell[l]{Cinq minutes après l'exposition, le vent commence à tourner, environ un minute plus tard, le vent atteint \\ 70 km/h, puis la pluie arrive, mais si forte et si grande qu'elle vous frappe la peau comme une aiguille, puis \\ le hail tombe du ciel, les gens paniquent, s'expriment et se courent l'un sur l'autre.} &   \textbf{0.72} & 39.90 \\ \hline
\end{tabular}
}
\end{table*}

\end{document}